\documentclass[
pdflatex,
sn-basic]{sn-jnl}

\usepackage[english]{babel}
\usepackage{csquotes}
\usepackage{natbib}
\usepackage{mathtools,amsmath,amssymb,amsfonts,dsfont}
\usepackage{siunitx}
\usepackage{enumitem}
\usepackage{graphicx}
\usepackage{calc}
\usepackage{pdflscape}
\usepackage{xcolor}
\usepackage{url}

\makeatletter

\usepackage{xspace} 

\newcommand*{\@appenddot}[1]{\@ifnextchar{.}{#1}{}#1.}
\newcommand*{\etc}{\@appenddot{etc}\@\xspace}
\newcommand*{\forexample}{\@appenddot{e.g}\@\xspace}
\newcommand*{\idest}{\@appenddot{i.e}\@\xspace}
\newcommand*{\etal}{et \@appenddot{al}\@\xspace}
\newcommand*{\cf}{\@appenddot{cf}\@\xspace}

\newcommand{\defn}[1]{\emph{#1}}
\newcommand*{\Concept}[1]{\textsf{#1}}
\newcommand*{\parhead}[1]{\textbf{\@appenddot{#1}}\@\xspace}
\newcommand{\CAmethod}[2]{%
  \par\noindent\parhead{#1}
}
\newcommand{\Proxymethod}[2]{\CAmethod{#1}{#2}}
\DeclareMathOperator*{\mean}{mean}
\newcommand{\rot}[2][80]{\rotatebox{#1}{\parbox{14ex}{\scriptsize\textbf{#2}}}}
\newcommand{\secseprule}[1]{\cmidrule[\heavyrulewidth]{1-#1}}
\newcommand{\methodtabsref}{Tables~\ref{tab:methods.enforcement.single}--\ref{tab:methods.posthoc.mining}}

\makeatother

\newlist{parlikedescription}{description}{1}
\setlist[parlikedescription]{
  font=\normalfont\parhead, leftmargin=0pt,
  parsep=.2\baselineskip,
}

\usepackage{hyperref}
\usepackage{booktabs,tabulary}
\usepackage{ltablex}

\jyear{2021}
\title[Concept Embedding Analysis]{Concept Embedding Analysis: A Review}

\author[1]{\fnm{Gesina} \sur{Schwalbe} (ORCID: 0000-0003-2690-2478)}
\email{gesina.schwalbe@continental-corporation.com}

\affil*[1]{
  \orgname{Continental AG},
  \orgaddress{
    \city{Regensburg},
    \country{Germany}}%
}

\hypersetup{
  pdfauthor={G. Schwalbe},
  pdftitle={Concept Embedding Analysis: A Review},
}

\addto\extrasenglish{%
}

\begin{document}


\abstract{
%
%
%

Deep neural networks (DNNs) have found their way into many applications
with potential impact on the safety, security, and fairness
of human-machine-systems. Such require basic
understanding and sufficient trust by the users.
This motivated the research field of explainable artificial
intelligence (XAI), \idest finding methods for opening the
\enquote{black-boxes} DNNs represent.
For the computer vision domain in specific,
practical assessment of DNNs requires a globally valid association
of human interpretable concepts with internals of the model.
The research field of \emph{concept (embedding) analysis (CA)}
tackles this problem:
CA aims to find global, assessable associations of
humanly interpretable semantic concepts (\forexample, \Concept{eye},
\Concept{bearded}) with internal representations of a DNN.
%
This work establishes a general definition of CA and a taxonomy for
CA methods, uniting several ideas from literature.
That allows to easy position and compare CA approaches.
Guided by the defined notions, the current state-of-the-art research regarding CA methods and
interesting applications are reviewed. More than thirty relevant
methods are discussed, compared, and categorized.
Finally, for practitioners, a survey of fifteen datasets is provided that
have been used for supervised concept analysis.
Open challenges and research directions are pointed out at the end.



  \keywords{concept analysis, concept activation vector, XAI, taxonomy, deep learning}
}

\maketitle

\bmhead{Acknowledgments}
The research leading to these results was partly funded by by the
German Federal Ministry for Economic Affairs and Energy within the
projects
\enquote{KI Wissen – Automotive AI powered by Knowledge}
and \enquote{KI Absicherung – Safe AI for automated driving}.
Thanks to the consortia for the successful cooperation.



\section{Introduction}
%
%
%

Deep neural networks have become a key to many
application fields. However, with their popularity,
their potential impact to safety and fairness increased
to an extend that the need for responsible artificial intelligence
\citep{arrieta_explainable_2020} became apparent.
For example, the European Union General Data Protection Regulation
enacted in 2016 demands algorithms not only to be efficient,
but also transparent and fair \citep{goodman_european_2017}.
Similarly, the automotive functional safety standard
ISO\,26262~\citep{iso/tc22/sc32_iso_2018} considers manual inspection
of algorithms an important measure of safety assurance.
These requirements gave rise to the field of explainable artificial
intelligence which was boosted in 2016 by the United States Defense
Advanced Research Projects Agency (DARPA)
\citep{darpa_explainable_2016,adadi_peeking_2018}.
XAI in general tries to translate behavioral or internal aspects
of black-box algorithms like DNNs into a human interpretable form
\citep{schwalbe_xai_2021}.

While many different aspects of a DNN can be the subject of a human
understandable explanation, their safety assessment requires
interpretable access to the \emph{internal representation}.
This requirement holds even for hardly specifiable tasks with
non-symbolic inputs like object detection from images.
This access allows to debug or audit the DNN function with
respect to failure modes that violate symbolic prior knowledge
(\forexample, \enquote{\Concept{eyes} usually belong to a
  \Concept{movable object}});
and this access potentially points out how to fix inappropriate
internal representations or directly enforce sufficient ones.
Besides the subject of explanation,
another relevant property of explainability methods is their locality
or scope \citep{adadi_peeking_2018}:
Many methods for analyzing computer vision DNNs produce local
explanations specific to one or few samples \citep{zhang_visual_2018}.
In contrast, \emph{global} explanations are valid for many more examples than would
be possible with manual inspection of many local explanations. This
makes global explanations interesting for practical assessment tasks.
Another practical constraint imposed by assessment use-cases is that
explanations must provide high \emph{fidelity}
and that they must be \emph{simple}
(\forexample, linear transformations of the object of explanation,
as suggested by \cite{kim_interpretability_2018})
to avoid introducing further complexity.
Lastly, assessments also require analysis of feature
interdependencies and interaction. This is another drawback of solely
using the popular heatmapping 
methods~\citep{zhang_examining_2018,janizek_explaining_2020} that
highlight local attention of a DNN, like
relevance~\citep{bach_pixelwise_2015},
saliency~\citep{sundararajan_axiomatic_2017}, or
perturbation~\citep{fong_interpretable_2017} based ones.

A promising class of methods fulfilling the mentioned desires is
that of \defn{concept (embedding) analysis (CA)} methods.
The main idea of concept embedding analysis is to associate
(in a simple way)
semantic concepts taken from natural human language, like \Concept{eye},
with vectors---the concept embedding or
\defn{concept activation vectors (CAV)}~\citep{kim_interpretability_2018}---,
or sub-spaces in the intermediate output space of one (or several)
layer(s) in a DNN, the latent space(s).
These approaches allow
global
access to latent space representations of semantic concepts in an
interpretable way,
without introducing further complexity.
Thus, they fulfill the mentioned desires for assessment tasks and
model improvement.

Early work on concept analysis started with unsupervised visualization
of single features in convolutional DNNs (CNNs), \idest neurons,
filters or groups thereof \citep{olah_feature_2017}.
Later on, supervised approaches were based on the NetDissect method
developed in \cite{bau_network_2017}, where single filters of CNNs
were associated with predefined visual semantic concepts.
Two cornerstones towards concept analysis research were established
by \cite{fong_net2vec_2018} and \cite{kim_interpretability_2018} in 2018.
Both methods associate predefined semantic concepts with vectors in
the latent spaces, thus post-hoc disentangling the representation of
such semantic concepts within DNN internal representations.
On this basis, distributed representations in unsupervised
settings were first considered in the
ACE~approach~\citep{ghorbani_automatic_2019} proposed in 2019.
Another notable development was that of concept bottleneck
models~\citep{losch_interpretability_2019,koh_concept_2020}:
Stepping away from post-hoc explainability, these include prior
knowledge about task-relevant concepts directly in the architecture of
a DNN. This is done with a bottleneck layer in which single nodes are
trained to correspond to semantic concepts.
These baseline approaches were extended and generalized in many
interesting ways in recent years.
Thus, time seems to be ready to take a closer look on the
developments in this young research area,
as will be done in this review.

\paragraph*{Contributions}
This paper shall shed light onto concept analysis as a practically
interesting and emerging sub-field of XAI, in order to foster research
in this direction.
Its main contributions towards this are as follows:
\begin{itemize}
\item 
  A formal, general \emph{definition} is provided capturing the
  core ideas uniting concept analysis methods, and differentiating it
  from similar research fields
  (cf.~\autoref{sec:ca}).
\item 
  In order to help researchers to find further research directions and
  position their work,
  a general \emph{taxonomy} for concept analysis approaches
  is developed and applied to state-of-the-art research methods
  (cf.~\autoref{sec:methods} and \autoref{fig:taxonomy}).
\item
  The breadth of the topic is demonstrated by an in-depth
  and systematic review of more than 30 diverse
  \emph{concept analysis approaches}.
  These are discussed, compared, and categorized according to the taxonomy
  (cf.~\autoref{sec:methods} and \methodtabsref).
\item
  An overview of more than 15 image \emph{datasets} used for concept
  analysis is compiled
  (cf.~\autoref{sec:datasets} and \autoref{tab:datasets}).
\item
  Based on the taxonomy and the method reviews,
  open challenges and promising \emph{research directions} are
  uncovered and discussed
  (cf.~\autoref{sec:challenges}).
\end{itemize}

\paragraph*{Scope}
This review goes along with the baseline work from
\cite{kim_interpretability_2018,fong_net2vec_2018,bau_network_2017,ghorbani_automatic_2019}.
The main focus lies on visual interpretability,
considering computer vision tasks as guiding examples.
The literature review combined a two-level recursive reference search
starting with the mentioned baselines, and a keyword search based on
the keywords
\enquote{Concept Analysis} and
\enquote{Concept Embedding Analysis}
on the search engine
Google Scholar\footnote{\url{https://scholar.google.com/}}.
Papers were included by a two-stage review approach, first filtering
by title and then by the abstract.
Included were global interpretability methods that pursue the mentioned
general goal of associating human interpretable concepts with DNN
latent space vectors or sub-spaces.

\paragraph*{Outline}
This paper is structured as follows:
In \autoref{sec:relatedwork}, related work and research fields are
recapitulated to demarcate the scope of concept analysis as considered
here.
A precise definition for concept analysis is then developed in
\autoref{sec:ca}, and a detailed taxonomy for concept analysis is
prescribed.
The breadth of the topic is demonstrated in
\autoref{sec:methods}, in which interesting concept analysis
methods are outlined, compared, and classified according to the derived
taxonomy (cf.\ \methodtabsref).
Applications of concept analysis are then
discussed in \autoref{sec:applications}, including model distillation
(cf.~\autoref{tab:proxytypes}) and qualitative as well as quantitative DNN
assessment methods.
Accompanying these collections, \autoref{sec:datasets}
compiles an overview on image datasets for supervised concept analysis
that are used in methods presented throughout this paper.
This may serve as a starting point and reference for researchers
looking for typical evaluation datasets.
Section~\ref{sec:challenges} provides an outline of current
challenges in the area of concept analysis, and an outlook on
potential research directions.


\section{Related work}\label{sec:relatedwork}
%
%
%

The field of XAI has seen an exponential rise in research interest in
the last decade
\citep{vilone_explainable_2020,arrieta_explainable_2020,linardatos_explainable_2021,zhou_evaluating_2021}.
By now, good introductory work is available, like \cite{molnar_interpretable_2020},
and many extensive method surveys,
including general ones like \cite{%
  linardatos_explainable_2021,%
  vilone_explainable_2020,%
  arrieta_explainable_2020,%
  carvalho_machine_2019,%
  adadi_peeking_2018,%
  gilpin_explaining_2018%
},
and ones specific to sub-topics like
reinforcement learning~\citep{heuillet_explainability_2021},
visual interpretability~\citep{tjoa_survey_2020,zhang_visual_2018}, or
rule extraction~\citep{hailesilassie_rule_2016}.
A meta-review for XAI methods can be found in \cite{schwalbe_xai_2021}.
We also consider the typical XAI method classification aspects of
black- versus white-box,
post-hoc versus inherent interpretability, and
simplicity.
These considerations are further refined and tailored to concept
analysis in specific, and extended by implementation specific
properties of CA methods.
A first review on concept analysis is provided by
\cite{kazhdan_disentanglement_2021}, with a two-class scheme for
classifying concept analysis methods. In contrast to our broad
review, the focus in that work lies on the general practical
comparison of two concept analysis approaches with latent space
disentanglement.
Especially, our review for the first time gives a broad overview of
the methods and applications of concept analysis,
providing a common point of view of methods
usually sorted into very different categories (\forexample, post-hoc
and inherent interpretability).
In the following, related XAI research fields are distinguished from
concept analysis.

\paragraph*{Visual attribution methods}
A very popular subfield of XAI is about providing
heatmaps that point out what input region
(\forexample, pixels in an image or words in a text)
the DNN mostly attended to make a concrete decision,
\idest which region had the highest \defn{attribution}.
Notable types are
perturbation based methods~\citep{ribeiro_why_2016,fong_interpretable_2017},
relevance back-propagation such as LRP~\citep{bach_pixelwise_2015},
activation map based approaches~\citep{zhou_learning_2016},
and gradient respectively sensitivity based
approaches~\citep{sundararajan_axiomatic_2017}.
All these approaches find their use in some of the
concept analysis applications presented here, \forexample, for
training~\citep{wu_global_2020},
more precise localization of concepts~\citep{lucieri_explaining_2020,fong_net2vec_2018}, or
dependency analysis~\citep{kim_interpretability_2018}.
While the heatmaps essentially provide a simple linear approximation
of the DNN behavior~\citep{ribeiro_why_2016}, these explanations are
inherently local and do not provide insights into internal encodings
of the DNN. This also prohibits deeper analysis of relations between
concepts \citep{rabold_expressive_2020}.
In contrast, the underlying mapping of concepts to internal features
of concept analysis methods is inherently global.

\paragraph*{Latent space disentanglement}
Concept analysis is very closely related to latent space
disentanglement~\citep{kazhdan_disentanglement_2021} methods like
VAEs~\citep{kingma_autoencoding_2014,higgins_betavae_2016},
as both approaches try to represent (parts of) the latent space as interpretable
features.
Similar to so-called concept bottleneck
models~\citep{koh_concept_2020}, disentanglement methods have a
bottleneck layer in which single units are trained to encode a
feature.
These features, however, are not necessarily aligned with human
semantics (especially when trained unsupervised),
but instead should encode the most prevalent independent factors of
variation causally related to the DNN output.
The target features of concept bottleneck models instead are
\emph{interpretable} factors of
variation~\citep{kazhdan_disentanglement_2021} perceived by humans.
These are not necessarily---and often are not---independent,
but allow to assess the dependencies of semantic concepts
encoded by the DNN.


\section{Concept embedding analysis}\label{sec:ca}
%
%
%

To get a clear understanding of the scope of concept analysis
considered here, related terms are defined in
\autoref{sec:ca.def}, and optional but desirable properties are
collected from literature.
Moreover, a taxonomy is provided in \autoref{sec:ca.taxonomy},
that allows a detailed analysis and classification of a use-case or
method for CA. The taxonomy is accompanied by illustrative examples,
and later applied to the sample methods highlighted in
\autoref{sec:methods} (cf.\ \methodtabsref).

\subsection{Definitions}\label{sec:ca.def}

The output of a neural network layer spans a vector space. In
this paper, this is called \defn{latent space} of the layer, or
\defn{activation maps} if collections of units spatially correspond to
patches in the input as in the case of convolutional layers.
In this work the term \defn{concept (embedding) analysis} generally
refers to investigation of the questions 
\emph{whether}, \emph{how well}, \emph{how}, and with \emph{which properties}
information about a \defn{semantic concept} is represented within the
latent spaces or sub-spaces thereof.

%
\paragraph*{Semantic concepts}
A \defn{(semantic) concept}, also called \defn{attribute} (attr.)
in \cite{kronenberger_dependency_2019}, is a concept occurring in a natural
language, \idest a notion that can be described using natural
language.\footnote{
  Concept terms are marked \Concept{like this} throughout this work.
}
Examples are the synonym sets established in
WordNet~\citep{fellbaum_wordnet_1998}.
The semantic concepts of interest here are those that can be assigned
to (patches of) a DNN input sample in the form of additional labels.
Hence, the selection of concepts depends on the context given by the task.
In \cite{kazhdan_disentanglement_2021} this dependency is emphasized
in the definition of concepts as interpretable factors of
variations for a given training dataset.
The paper by \cite{bau_network_2017} introduced the following classes of semantic
concepts for images:
sample-level attributes like scene (\forexample, \Concept{rainy}) and
image qualities~\citep{abid_meaningfully_2021} (\forexample, \Concept{contrast});
full objects (\forexample, \Concept{person});
part objects (\forexample, \Concept{head}, \Concept{leg});
and object-level attributes like
material (\forexample, \Concept{wooden}),
texture (\forexample, \Concept{striped}), and
color (\forexample, \Concept{green}).
Other examples of object-level attributes are \Concept{gender},
and facial attributes like \Concept{bearded}
as considered in \cite{kim_interpretability_2018}.
For supervised concept analysis, it is assumed that labeled examples
of the concepts of interest are available, and
different approaches require different label formats.
Formats considered here are the most common
binary labels
(\forexample, \cite{kim_interpretability_2018}) 
respectively binary segmentation masks
(\forexample, \cite{fong_net2vec_2018,schwalbe_verification_2021}),
multi-class concepts \citep{kazhdan_now_2020},
and regression concepts
\citep{graziani_regression_2018,graziani_concept_2020}.
Other possible types are multi-valued concepts with independent
values (\forexample, \Concept{weather} with \Concept{raininess} and
\Concept{cloudiness}),
and spatial allocation of concepts, \forexample, via bounding boxes or
x-y-positions.
Note that the direct output of the DNN usually also relates to a
semantic concept, like objects for object detectors.
The primary interests here are concepts that are not included as
features in the DNN final output, \forexample, body parts in the case
of a pedestrian detector.

\paragraph*{Concept analysis (CA)}
By basic \defn{concept analysis} or concept embedding analysis we
define any activity that tries to associate semantic concepts with
vectors or sub-spaces in one preselected latent space of a DNN.
Further, we assume that this is done via a simple, global model
that allows to predict information about the concept from the latent
space activations of an input.
The model is called the \defn{concept model} of the concept,
and its output depends on the label type of the concept (\forexample,
binary presence of the concept, localization, regression,
classification, etc.).
\defn{Simple} means transparent or human interpretable, or not
introducing much more non-interpretable complexity,
like linear models~\citep{kim_interpretability_2018}.
The pair of a concept and its latent space representation (vector or
sub-space) is called the \defn{concept embedding}.
In case a vector is associated, this vector is called the
concept vector or \defn{concept activation vector
  (CAV)}~\citep{kim_interpretability_2018}.

\paragraph*{Concept analysis goals}
As mentioned previously, the main goals of CA are to answer
\emph{whether}, \emph{how well}, \emph{how}, and with \emph{which properties}
information about a predefined or mined semantic concept is
represented within a latent space.
To just show \emph{whether} information about a given semantic concept
is available within a latent space, one can either perform a supervised
training of a concept model and check whether sufficient performance
can be reached~\citep{schwalbe_verification_2021},
as illustrated in \cite{fuchs_neural_2018};
or perform an unsupervised concept mining, later comparing found concept
associations with the given concept.
%
A proxy to measure \emph{how well} the information is encoded
generally is the concept model prediction performance.
%
The assessment of properties now requires finding an exact
representation of the concept information that can be related or even
compared to representations of other concepts (including the input and
output).
Most common are concept representations by CAVs.
The general idea of CAVs is to find a vector which is close to
the latent space vectors of examples of the given concept with respect
to some proximity measure.

\paragraph*{Desirable properties of concept models}
The following properties of concept models were considered useful in
literature:
\begin{itemize}
\item
  \emph{Comparable} concept models allow to assess whether the mapping
  between semantic and latent space preserves semantic similarities.
  In other words, whether semantically similar concepts are assigned to similar
  latent space representations~\citep{schwalbe_verification_2021}.
\item
  \emph{Vector representations} allow to utilize the rich vector space
  structure of latent spaces, including natural comparability:
  \begin{itemize}
  \item 
    In \cite{mikolov_linguistic_2013} it was found that for
    continuous space language models the \emph{cosine similarity} on
    latent space vectors correspond to meaningful semantic
    (cor-)relations of the encoded words. An example is the famous
    relation
    \enquote{$\text{\Concept{king}} - \text{\Concept{man}} +
      \text{\Concept{woman}} = \text{\Concept{queen}}$}.
    Net2Vec~\citep{fong_net2vec_2018} and
    TCAV~\citep{kim_interpretability_2018} showed that this also holds for
    CAVs in computer vision networks,
    and \cite{zhang_examining_2018} used cosine similarity for local
    CAVs modeled by masked derivative vectors.
  \item
    The local \emph{sensitivity} of later layer representations with respect
    to the concept can easily be measured via the partial derivative
    along the CAV as proposed in \cite{kim_interpretability_2018}.
  \end{itemize}
\item
  \emph{Linear concept models} are considered to be natural and easy to
  assess for humans in \cite{kim_interpretability_2018}.
\item
  \emph{Sparsity} of CAVs in case of linear encodings is postulated to
  increase interpretability of the concept
  representations~\citep{bau_network_2017}.
  \cite{bau_network_2017} even suggest using the
  number of concepts to which a CNN filter contributes
  as a comparative measure for interpretability or entanglement---the
  more filters, the less interpretable.
\item
  Concept models allowing \emph{gradient back-propagation}, like
  Net2Vec, can be utilized as additional
  outputs during training or \emph{fine-tuning}~\citep{schwalbe_concept_2020}.
  Also, they allow to measure \emph{attribution} of earlier
  layer concepts or inputs to a given concept output via relevance
  back-propagation or sensitivity~\citep{lucieri_explaining_2020}.
  Otherwise, one has to turn to perturbation-based attribution
  methods~\citep{wang_chain_2020}.
\end{itemize}


\subsection{Taxonomy}\label{sec:ca.taxonomy}

\begin{figure}
  \centering
  \hspace*{-.05\linewidth}\includegraphics[width=1.1\linewidth]{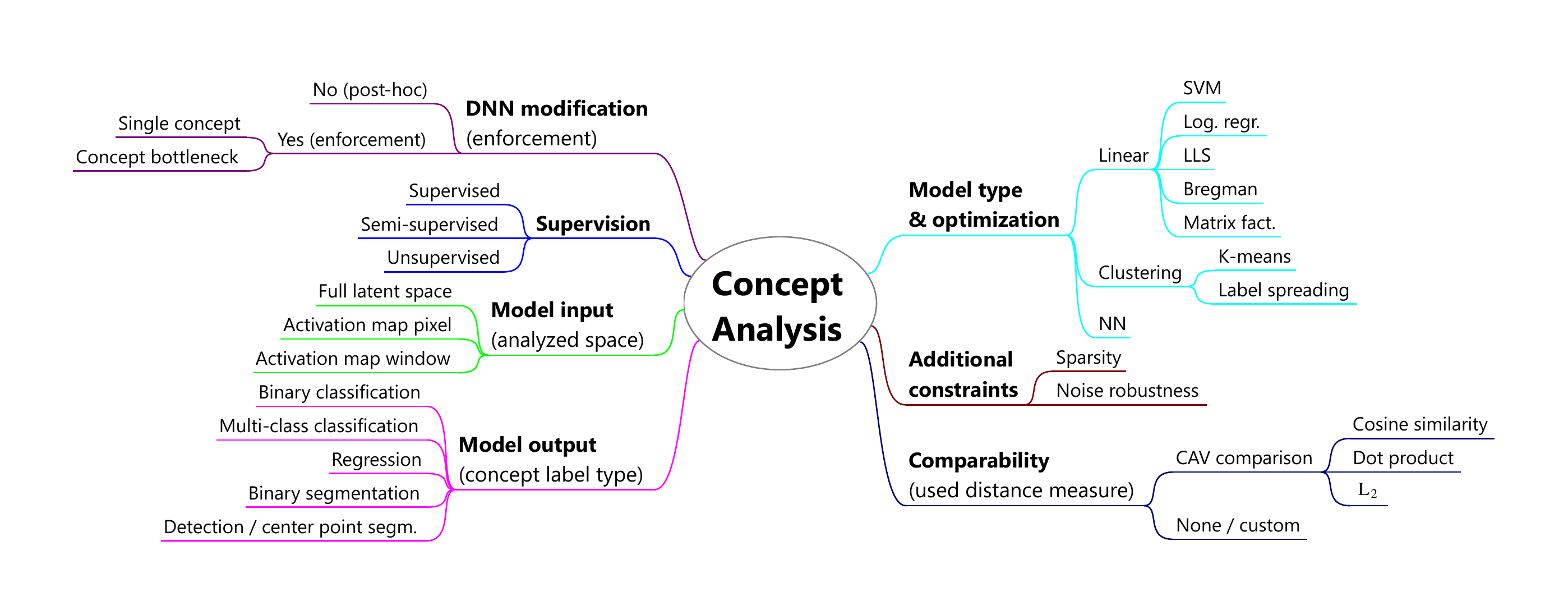}
  \caption{
    Overview on the discussed taxonomy aspects for concept analysis
  }
  \label{fig:taxonomy}
\end{figure}

In the following we collect key aspects to differentiate approaches
for concept embedding analysis, summarized in \autoref{fig:taxonomy}.
An overview of how the methods discussed in \autoref{sec:ca} fit into
the suggested categorization scheme can be found in \methodtabsref.

\paragraph*{Concept model supervision}
The concepts to be analyzed can either be predefined by sample data and
then associated to DNN internal representations in a supervised (superv.) fashion, or
human interpretable concepts can be \emph{mined} from the latent spaces
in an unsupervised way.
One example of the latter is Automated Concept-based Explanations
(ACE) by \cite{ghorbani_automatic_2019}.
They search for clusters in the latent space vectors of super-pixels
(the concept candidates),
defining a concept by a cluster and the concept vectors as the cluster
center.
Another example of unsupervised concept analysis is feature
visualization~\citep{olah_feature_2017}.

\paragraph*{Concept model output: Concept label type}
First concept analysis approaches considered binary concept model
outputs, either binary classification (cls) labels as in
TCAV~\citep{kim_interpretability_2018} and ACE,
binary segmentation (seg) masks as in Net2Vec~\citep{fong_net2vec_2018},
or binary concept center point prediction (det) masks as in
\cite{schwalbe_concept_2020,rabold_expressive_2020}.
This binary setting can be extended to multi-class (mcls) concepts like
\Concept{shape} as suggested in \cite{kazhdan_now_2020},
or a multi-dimensional concept like in the
IIN~\citep{esser_disentangling_2020} method, where concepts are
represented by sub-spaces of the transformed latent space.
Regression Concept Vectors (RCV) by \cite{graziani_regression_2018},
and its successor methods consider continuous-valued
instead of discrete-valued concepts.

\paragraph*{Concept model input}
TCAV considers the complete latent
space of a layer for its binary classification.
The same holds for successor methods that try to mitigate the
resulting high memory demand by average pooling of the latent space,
like \cite{graziani_concept_2020,chyung_extracting_2019}.
However, it is hard to localize concepts in the input only from image
level concept information, even using attention
methods~\citep{lucieri_explaining_2020}.
To enable localization of concepts, Net2Vec
does not consider the complete activation space of a CNN,
but instead operates on single pixels of an activation map, utilizing
the spatial correspondence between activation map pixels and input
regions. In this case of a linear concept model, this can efficiently
be implemented by a $1\times1$-convolution.
The convolutional Net2Vec-approach is adapted in
\cite{schwalbe_concept_2020}, only using windows larger than
$1\times1$ pixels to provide enough context information as input to
the concept model.
The prior work of Net2Vec on network
dissection~\citep{bau_network_2017} was restricted not only to
$1\times1$-windows, but also to unit vectors
(\idest single filters) per activation map pixel,
similar to standard feature visualization~\citep{olah_feature_2017}.

%
\paragraph*{Concept model type and optimization}
The optimal retrieval of concept information from latent space input
can be solved via different models,
optimization methods, and
regularization constraints.
Examples for model types are:
small neural networks
(\forexample, Neural Stethoscopes by \cite{fuchs_neural_2018}, IIN);
k-means clustering
(\forexample, SeVec by \cite{gu_semantics_2019}, ACE);
or more general matrix factorization
(\forexample, NCAV by \cite{zhang_invertible_2021}; \cite{saini_analyzing_2020});
and linear models
(\forexample, TCAV, Net2Vec).
The latter features, \forexample, usage of
support vector machines (\forexample, TCAV),
logistic regression (log.\ regr.) (\forexample, Net2Vec, \cite{schwalbe_concept_2020}),
and Bregman iteration (\forexample, CHAIN by \cite{wang_chain_2020}) for optimization.
It also makes a difference whether a bias is allowed in the linear model or not.
Examples of regularization constraints are:
Sparsity, as aimed for in NetDissect~\citep{bau_network_2017}, and in
CHAIN via $L_1$ loss; and
robustness against noise as implemented in
Net2Vec and
NetDissect by thresholding activation maps.

\paragraph*{Enforcement or post-hoc}
As for other XAI methods, CA approaches can also be classified into
inherent interpretability methods enforcing a level of transparency,
and post-hoc methods not changing the analyzed DNN like
\cite{fong_net2vec_2018,kim_interpretability_2018}.
Supervised post-hoc CA is called \defn{concept localization},
and unsupervised post-hoc CA is referred to as \defn{concept mining}. 
Typical inherently interpretable architectures utilizing CA are
so-called \defn{concept-bottleneck models (CBM)}:
These DNNs feature a bottleneck layer in which all or a sub-set of the
nodes are trained to be directly associated with a semantic concept.
Example architectures and training schemes can be found in
\cite{koh_concept_2020,losch_interpretability_2019} (with concept
supervision), and
ProtoPNet~\citep{chen_this_2019} (without concept model output
supervision).
Different architectures to integrate and connect the concept
bottleneck into the DNN are compared in \cite{kronenberger_dependency_2019}.
Interestingly, good embeddings of task relevant concepts seem to
support or at least not decrease the final performance of the network
\citep{losch_interpretability_2019}, and allow for
better plausibility checks during operation \citep{kronenberger_dependency_2019}.

\paragraph*{Comparability: CAV distance measure}
In case concept activation vectors are used to represent concepts in
the latent space, both finding and interpreting the vectors often
relies on a vector distance measure:
For a concept, a CAV is selected which is \enquote{close} to latent
space vector representations of samples; and to compare concept models
based on CAVs, one can also utilize this proximity measure.
To our knowledge, three measures have been used in literature:
\begin{itemize}
\item \emph{$L_2$ distance}
  is used for concept mining via clustering in ACE
  and its successor \cite{yeh_completenessaware_2020}.
  It is also used for training a concept localization linear model in
  the method CHAIN,
  and the matrix factorization approach NCAV.
\item \emph{Cosine similarity}
  between two vectors $v_1, v_2$ encodes the angle
  between the vectors and is defined as the normalized dot product
  \begin{gather*}
    \text{CosSim}(v_1, v_2) \coloneqq
    \frac{v_1\circ v_2}{\|v_1\|\cdot\|v_2\|} \in [0,1]
  \end{gather*}
  which is 1 for parallel, 0 for orthogonal, and -1 for anti-parallel
  vectors.
  It was used in SeVec~\citep{gu_semantics_2019} for concept
  localization via clustering.
\item \emph{Dot product}
  is the standard distance measure used to define linear models:
  The model defines a hyperplane
  $H = \{ d_H(v) = v \mid v\circ v_H + b_H = 0\}$
  (biased by $b_H$)
  where $v_H$ is the normal vector (the concept vector) and
  $d_H$ defines the signed shortest distance to $H$.
  This is utilized in the concept localization approaches of
  TCAV,
  Net2Vec, and their successors
  \cite{schwalbe_concept_2020,schwalbe_verification_2021}, and
  \cite{kronenberger_dependency_2019}.
\end{itemize}
Other known model types have to use learned distance metrics for
comparability.


\section{Concept analysis methods}\label{sec:methods}
%
%
%

As becomes apparent from the count of taxonomy aspects, there is an
abundance of methods to obtain the desired additional concept outputs
and concept models inherent to concept analysis.
This section deals with prominent example methods to achieve this, and
categorizes them according to main aspects of the proposed taxonomy.
The first part in \autoref{sec:methods.enforcement} discusses
methods that work towards inherent transparency and change the
underlying model architecture.
And the second part in \autoref{sec:methods.posthoc}
presents methods working on pre-trained models without
modifying them, \idest post-hoc explainability methods.
An overview over discussed methods is given in \methodtabsref.
Applications for the concept models and outputs will be discussed
later in \autoref{sec:applications}.

\renewcommand{\rot}[2][70]{\rotatebox{#1}{\parbox{10ex}{\footnotesize\textbf{#2}}}}
\newcommand{\methodsecseprule}{\secseprule{8}}
\newcommand*{\methodcite}[1]{{\footnotesize \mbox{\citep{#1}}}}

\newcommand{\methodtabcontent}[3]{%
  \expandafter\gdef\csname methodtabcontent.#1\endcsname{%
    \methodsecseprule\multicolumn{8}{@{}l@{}}{\textbf{#2 (\autoref{sec:#1})}}\\*\midrule%
    #3%
  }%
}

\newcommand{\methodtab}[1]{%
  \begin{table}
    \centering
    \caption{Properties of concept analysis
      methods discussed in \autoref{sec:#1}
      according to the taxonomy defined in \autoref{sec:ca.taxonomy}.}
    \label{tab:#1}
    \begin{tabulary}{\linewidth}{@{}>{\raggedright}p{15em} @{}c@{}c@{\;} l l@{\;} >{\raggedright}p{\widthof{\footnotesize spreading}} l@{\;} L@{}}
      \methodtabheader{}\\%
      \csname methodtabcontent.#1\endcsname%
      \\\bottomrule
    \end{tabulary}
  \end{table}
}

\newcommand{\methodtabheader}{%
\rot[0]{\scriptsize Method}	&	\rot{Enforced}	&	\rot{Superv.}	&	\rot{Model}	&	\rot{Distance measure}	&	\rot{Optim.}	&	\rot{Label type}	&	\rot{Input}	
}
\methodtabcontent{methods.enforcement.single}{Single concept enforcement}{																	
Neural Stethoscopes \methodcite{fuchs_neural_2018}	&	\checkmark	&	\checkmark	&	DNN	&	learned	&	log.\,regr.	&	cls	&	full	\\\midrule	
EDD \methodcite{kronenberger_dependency_2019}	&	(\checkmark)	&	\checkmark	&	linear	&	dot	&	log.\,regr.	&	cls	&	full	\\\midrule	%
UPerNet \methodcite{xiao_unified_2018}	&	\checkmark	&	\checkmark	&	DNN	&	learned	&	log.\,regr.	&	seg	&	window	\\\midrule	
Interpretable CNNs \methodcite{zhang_interpretable_2018}	&	\checkmark	&	-	&	linear	&	dot	&	--	&	cls	&	window		
}%
\methodtabcontent{methods.enforcement.cbm}{Concept bottleneck models}{																	
Concept Whitening \methodcite{chen_concept_2020}	&	\checkmark	&	\checkmark	&	linear	&	dot	&	LPQC	&	any	&	full	\\\midrule	%
ProtoPNet \methodcite{chen_this_2019}, CSPP \methodcite{feifel_reevaluating_2021}	&	\checkmark	&	-	&	cluster	&	$L_2$	&	custom	&	det	&	window	\\\midrule	
Semantic bottlenecks \methodcite{losch_interpretability_2019}	&	\checkmark	&	\checkmark	&	linear	&	dot	&	log.\,regr.	&	seg	&	pixel	\\\midrule	
Concept bottlenecks \methodcite{koh_concept_2020}	&	\checkmark	&	\checkmark	&	linear	&	dot	&	log.\,regr.	&	cls	&	full	\\\midrule	
Weakly supervised CBMs \methodcite{belem_weakly_2021}	&	\checkmark	&	\checkmark	&	linear	&	dot	&	log.\,regr.	&	cls	&	full	\\\midrule	
Concept Groups \methodcite{marcos_contextual_2020}	&	\checkmark	&	\checkmark	&	custom	&	dot	&	custom	&	cls	&	full		
}%
\methodtabcontent{methods.posthoc.localization}{Post-hoc concept localization}{																	
NetDissect \methodcite{bau_network_2017}	&	-	&	\checkmark	&	linear	&	dot	&	picking	&	seg	&	pixel	\\\midrule	
Net2Vec \methodcite{fong_net2vec_2018}{\footnotesize{},} \cite{schwalbe_verification_2021}	&	-	&	\checkmark	&	linear	&	dot	&	log.\,regr.	&	seg	&	pixel	\\\midrule	
\cite{schwalbe_concept_2020}, \cite{rabold_expressive_2020}	&	-	&	\checkmark	&	linear	&	dot	&	log.\,regr.	&	det	&	window	\\\midrule	%
TCAV \methodcite{kim_interpretability_2018}, \cite{chyung_extracting_2019}	&	-	&	\checkmark	&	linear	&	dot	&	SVM	&	cls	&	full	\\\midrule	
CLM \methodcite{lucieri_explaining_2020}	&	-	&	\checkmark	&	linear	&	dot	&	SVM	&	seg	&	full	\\\methodsecseprule	
RCV \methodcite{graziani_regression_2018,graziani_concept_2020}	&	-	&	\checkmark	&	linear	&	dot	&	LLS	&	reg	&	full	\\\midrule	
CHAIN \methodcite{wang_chain_2020}	&	-	&	\checkmark	&	linear	&	$L_2$	&	Bregman	&	seg	&	pixel	\\\midrule	
Local CAVs \methodcite{zhang_examining_2018}	&	-	&	\checkmark	&	loc.\,lin.	&	cosine	&	Lasso	&	seg	&	window	\\\midrule	
AfI \methodcite{wu_global_2020}	&	-	&	\checkmark	&	loc.\,lin.	&	dot	&	custom	&	cls	&	full	\\\methodsecseprule	
SeVec \methodcite{gu_semantics_2019}	&	-	&	\checkmark	&	cluster	&	cosine	&	?	&	cls	&	full	\\\midrule	
CME \methodcite{kazhdan_now_2020}	&	-	&	semi	&	cluster	&	learned	&	label \mbox{spreading}	&	mcls	&	full	\\\midrule	
IIN \methodcite{esser_disentangling_2020}	&	-	&	(\checkmark)	&	DNN	&	$L_2$	&	custom	&	any	&	full	\\\midrule	%
CBM student \methodcite{haselhoff_blackbox_2021}	&	-	&	\checkmark	&	linear	&	cosine	&	custom	&	cls	&	full		
}%
\methodtabcontent{methods.posthoc.mining}{Post-hoc concept mining}{																	
Feature visualization \methodcite{olah_feature_2017,nguyen_understanding_2019}	&	-	&	-	&	linear	&	$L_2$	&	--	&	seg	&	channel	\\\midrule	
ACE \methodcite{ghorbani_automatic_2019}, VRX \methodcite{ge_peek_2021}	&	-	&	-	&	cluster	&	$L_2$	&	k-means	&	cls	&	full	\\\midrule	%
\cite{yeh_completenessaware_2020}	&	-	&	-	&	cluster	&	$L_2$	&	custom	&	cls	&	full	\\\midrule	
NBDT \methodcite{wan_nbdt_2020}	&	(\checkmark)	&	-	&	cluster	&	$L_2$	&	HAC	&	cls	&	full	\\\midrule	
Explanatory Graph \methodcite{zhang_interpreting_2018}	&	-	&	-	&	(cluster)	&	\,-	&	custom	&	det	&	window	\\\methodsecseprule	%
NCAV \methodcite{zhang_invertible_2021},\newline\cite{saini_analyzing_2020}	&	-	&	-	&	linear	&	$L_2$	&	matrix fact.	&	seg	&	pixel		
%
}%


\subsection{Inherent concept models}\label{sec:methods.enforcement}
In many applications, inherent interpretability is desirable and may even 
improve model performance.
Examples are safety critical domains like automated driving or medicine,
or applications with ethical implications like job application preprocessing.
Structuring and training a DNN to use concept embeddings leads to more
interpretable intermediate representations respectively models.
In the following, it is differentiated between two types of enforcing
rich concept outputs:
(independent) enforcement of single concepts (\autoref{sec:methods.enforcement.single}),
and so-called concept bottleneck models (CBM) enforcing a complete
interpretable layer (\autoref{sec:methods.enforcement.cbm}).

\subsubsection{Single concept enforcement}\label{sec:methods.enforcement.single}
\methodtab{methods.enforcement.single}
One way to ensure good embeddings of concepts is to add additional outputs
for the respective concepts to the DNN and include their predictions into
the loss. Such context aware models can be structured like regular DNNs
but they use additional regularization to force the network to focus
on specific concept embeddings.
For example, \cite{schwalbe_concept_2020} suggested to attach a single neuron per
concept (respectively per concept location in the image) and simply add a
negative log-likelihood for the prediction performance as weighted summand to the loss.
\CAmethod{Neural Stethoscopes}{\citep{fuchs_neural_2018}}%
This was implemented in \cite{fuchs_neural_2018}. They attach 1-hidden-layer DNNs,
called Neural Stethoscopes, to the main model which are trained to predict an
image-level concept alongside the main output.
Their focus was on comparability of the layers, and the Neural
Stethoscopes can both be used for concept enforcement and post-hoc
concept analysis.
\CAmethod{EDD}{\citep{kronenberger_dependency_2019}}%
Similarly, Explanation Dependency Decomposition
(EDD) by \cite{kronenberger_dependency_2019} 
uses additional classifiers to extract the presence of visual
embeddings from the convolutional layer outputs.
They assessed diverse information flow settings: in-between concept models,
and between concept and main model outputs (for details
cf.~\cite[Fig.~2]{kronenberger_dependency_2019}).
On a simple traffic sign classification setup,
adding concept outputs did not infringe performance, and top
performance could be achieved by an architecture merging latent space
and concept model outputs before predicting the final output.
\CAmethod{UPerNet}{\citep{xiao_unified_2018}}%
This is extended to a new discipline, unified perceptual parsing,
in \cite{xiao_unified_2018}: The output of a DNN is not
restricted to a main task but extended by a large set of related concepts,
as demonstrated in their UPerNet architecture.
\CAmethod{Interpretable CNNs}{\citep{zhang_interpretable_2018}}%
While the other approaches are supervised and rely on concept labels,
\cite{zhang_interpretable_2018} enforces in an unsupervised fashion
the alignment of convolutional filters with concepts, which are in this case object parts.
This is done by using filter-specific losses that encourage the filter to
only activate locally around its maximum activation in the image.
This assumes that each object part type occurs once in an input image.

\subsubsection{Concept bottleneck models}\label{sec:methods.enforcement.cbm}
\methodtab{methods.enforcement.cbm}

\begin{figure}
  \centering
  \includegraphics[width=\linewidth]{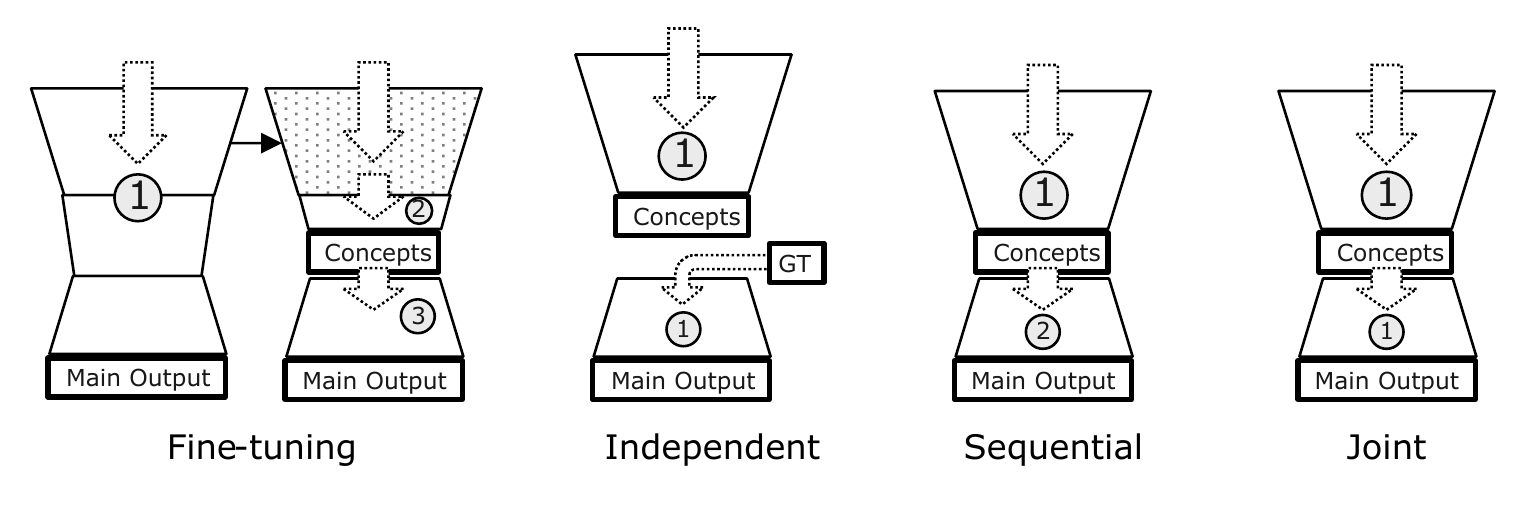}
  \caption{
    Different schemes for training concept bottleneck models both with
    the concepts and the main output objective,
    as suggested in literature
    \citep{losch_interpretability_2019,koh_concept_2020}
  }
  \label{fig:conceptbottlenecks}
\end{figure}

General concept enforcement will ensure a set of interpretable
outputs. So-called concept bottleneck models (CBM) additionally
require that the output of a complete layer should be interpretable,
\idest all units should correspond to human interpretable concepts.
CBMs primarily can be differentiated by the applied training scheme
(important ones depicted in \autoref{fig:conceptbottlenecks}).
%
\CAmethod{Concept whitening}{\citep{chen_concept_2020}}%
An example of a fine-tuning approach that re-trains a pretrained
network is concept whitening~\citep{chen_concept_2020}.
Concept whitening suggests a replacement layer for batch normalization
that is trained in a multi-task setting to apply batch whitening and
align the dimensions of the latent space to given concepts. This is
formulated as a linear programming problem with quadratic constraints
(LPQC) and requires few epochs to disentangle the dimensions after
replacing the batch normalization.
\CAmethod{ProtoPNet}{\citep{chen_this_2019}}
A clustering-based and unsupervised concept model approach is
considered by \cite{chen_this_2019} in form of the ProtoPNet architecture.
Here, the bottleneck layer consists of a prototype predictor:
Windows in the CNN layer activation output are compared to
unsupervised learned concept prototype CAVs, and the final output is
derived from the resulting prototype scores. The CAV cluster centers
and the rest of the model are trained jointly.
\CAmethod{CSPP}{\citep{feifel_reevaluating_2021}}
An extension to object detection tasks was proposed with the
Center Scale and Prototype Prediction
(CSPP) architecture by \cite{feifel_reevaluating_2021}.
\CAmethod{Semantic bottlenecks}{\citep{losch_interpretability_2019}}
The semantic bottlenecks architecture
\citep{losch_interpretability_2019}
also considers a fine-tuning training scheme.
Here, the output of a complete layer of a pretrained DNN is linearly
transformed to a representation where every filter represents one
predefined concept, while the later layers are retrained.
\CAmethod{CBM}{\citep{koh_concept_2020}}
Different schemes for directly training the concept bottleneck model
with both the concept and the main objective were compared in
\cite{koh_concept_2020}.
Settings were considered in which the part up-to the bottleneck and
the part top-of the bottleneck were trained
independently, jointly, or sequentially,
as illustrated in \autoref{fig:conceptbottlenecks}.
Results revealed joint training as the tight winner in performance.
As in \cite{losch_interpretability_2019}, performance was shown to be
competitive to non-interpretable standard models without concept
bottleneck, and also substantially more robust to spurious
correlations with background features.
\CAmethod{Weakly supervised CBM}{\citep{belem_weakly_2021}}%
Though \cite{koh_concept_2020} found that CBMs can mediocrely cope
with small data tasks, especially the joint training requires a
considerable amount of possibly costly concept labels.
Therefore, \cite{belem_weakly_2021} suggested a weak supervision of
concept models by generating noisy labels from approximate rules when
such are available from domain knowledge.
In a fraud detection example setting, they found sequential training
to work best, where first the noisy then the accurate labels are used.
\CAmethod{Concept Groups}{\citep{marcos_contextual_2020}}%
Another shortcoming of vanilla CBMs is that it may not be clear how a
predefined concept from the bottleneck is used for the final output,
respectively what its task-specific meaning is. To tackle this,
\cite{marcos_contextual_2020} suggests Concept Groups, a second
bottleneck layer directly attached to the first one where nodes
represent task-specific groups of concepts.
These are obtained unsupervised, and the complete model is trained in
a three-step sequential manner.

Lastly, it must be noted that the constraint imposed by a CBM,
namely that single units in a layer correspond to interpretable
concepts, does in general not guarantee that only information about
those concepts is passed through the layer.
Due to correlation of the concepts, further information can
be encoded which is possibly non-semantic, as was shown and criticized
in \cite{mahinpei_promises_2021}.


\subsection{Post-hoc concept models}\label{sec:methods.posthoc}
Many applications require working with a pretrained DNN without
changing its architecture. In such cases, post-hoc concept analysis
methods are required that train concept models on given latent space
outputs.
The two main classes of methods to associate single concepts with
latent space representations are supervised (concept
localization) and unsupervised (concept mining) post-hoc CA.
Furthermore, we will discuss examples of the special case when a
complete latent space is post-hoc disentangled on the basis of
semantic concepts.

\begin{figure}
  \centering
  \includegraphics[width=.4\linewidth]{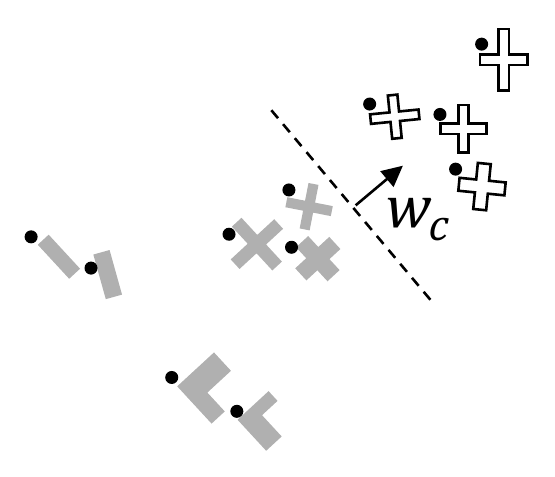}%
  \hfill%
  \includegraphics[width=.4\linewidth]{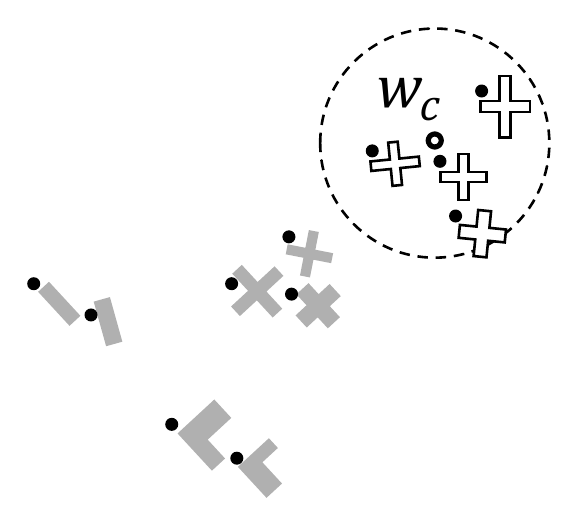}
  \caption{
    Comparison of concept vectors $w_c$ for two standard concept analysis
    model types, and the binary visual shape concept \Concept{plus} (unfilled).
    Dots ($\bullet$) represent points in the simplified latent space,
    accompanied by visualizations of the shape occurring in the image
    (patch) that produced this point.
    \emph{Left:}
    Linear concept model with a hyperplane as decision boundary and
    the hyperplane normal vector as CAV $w_c$;
    \emph{right:}
    $L_2$-clustering-based concept model with a spherical decision
    boundary and cluster center (unfilled) as CAV $w_c$.
  }
  \label{fig:camodelcomparison}
\end{figure}

\subsubsection{Concept localization}\label{sec:methods.posthoc.localization}
\methodtab{methods.posthoc.localization}
By now there is a formidable selection of supervised post-hoc concept
analysis methods described in literature.
For an overview, we cluster them by the model type into linear
and non-linear models.

\begin{figure}
  \centering
  \parbox[t]{.5\linewidth}{~\\[-\baselineskip]%
    \includegraphics[width=\linewidth]{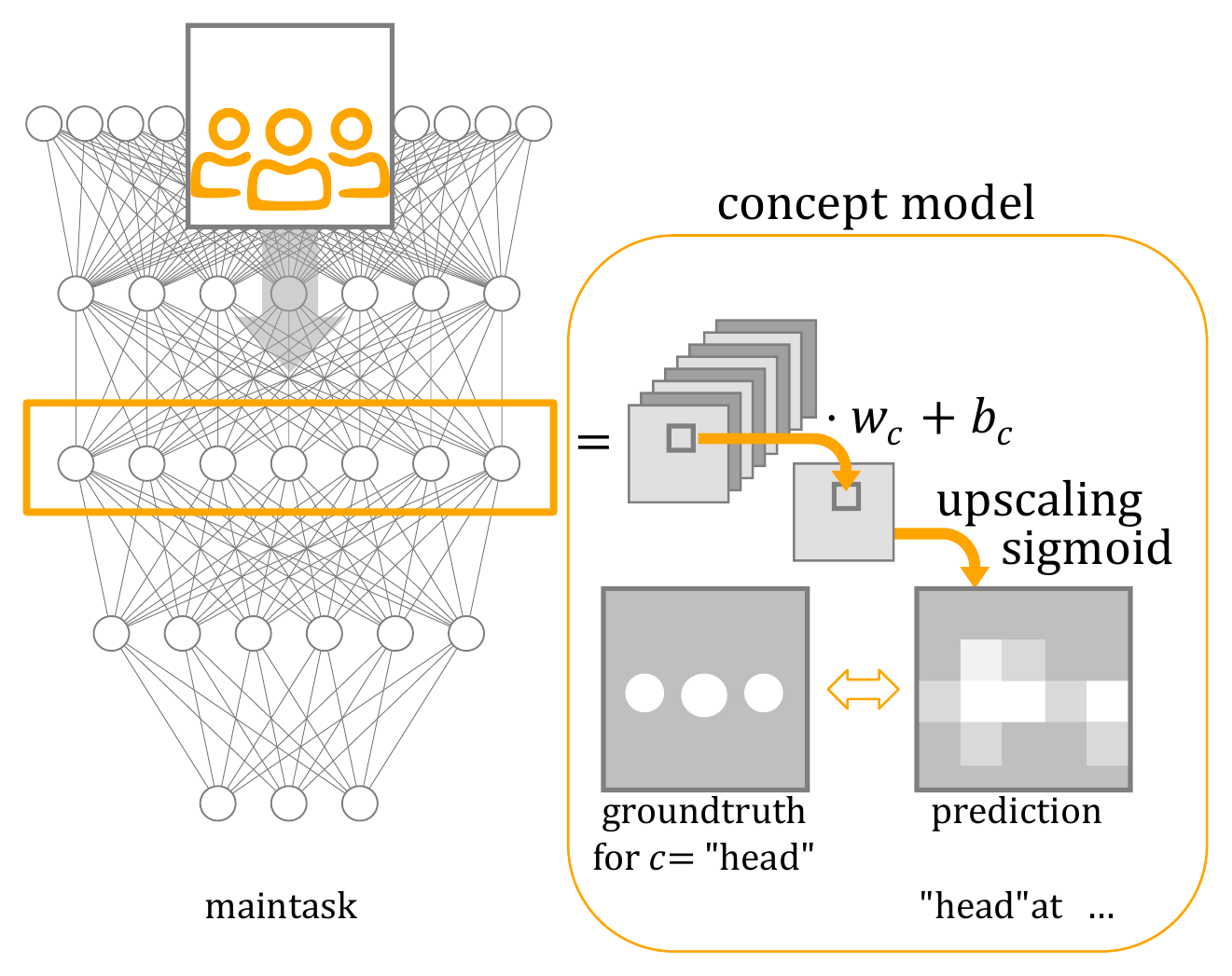}%
  }\hfill%
  \parbox[t]{.45\linewidth}{~\\[-\baselineskip]%
    \caption{
      Exemplary linear concept localization model from
      \cite{schwalbe_verification_2021}.
      The simple Net2Vec~\citep{fong_net2vec_2018} approach learns to
      predict segmentation masks for \Concept{head} using a 
      $1\times1$-convolution followed by upscaling and normalization.
      The kernel vector $w_c$ then is the CAV.
      \emph{(Figure taken from \cite[Fig.~1]{schwalbe_verification_2021})}
    }
    \label{fig:calinear}
  }
\end{figure}

\paragraph{Linear concept localization}
\CAmethod{NetDissect}{\citep{bau_network_2017}}%
An early work on supervised post-hoc concept analysis, and basis
for many further methods, was Network Dissection (NetDissect) by
\cite{bau_network_2017}. They associate single filters of a
convolutional DNN with one or several predefined concepts. To do
this, upscaled and denoised activation maps of each filter were
compared with ground truth concept segmentation masks, and the
concepts with best agreement are picked.
They also introduced the BRODEN dataset combination later often used
as a baseline in other methods.
\CAmethod{Net2Vec}{\citep{fong_net2vec_2018}}
A direct successor of NetDissect was Net2Vec~\citep{fong_net2vec_2018}
where not a filter is associated to several concepts, but a concept is
associated to a linear combination of filters. This is done by
training a linear model to classify denoised activation map pixels into
\enquote{belongs to concept} or \enquote{does not belong to concept},
resulting in binary segmentation masks.
For the implementation (cf.~\autoref{fig:calinear}),
a $1\times1$-convolution is used, the kernel
weights of which are the CAV for the concept.
For linear models like this the CAV points in the latent space direction
of the concept as illustrated in \autoref{fig:camodelcomparison}.
Investigation of those CAVs for the first time revealed that such
linearly obtained concept vectors behave like word vectors in word
vector spaces, with cosine similarity encoding some semantic
similarity.
\CAmethod{Net2Vec extensions}{\citep{schwalbe_concept_2020,rabold_expressive_2020}}%
There are several extensions of Net2Vec to concept center point
prediction~\citep{schwalbe_concept_2020,rabold_expressive_2020} instead
of binary segmentation, and to object
detection DNNs~\citep{schwalbe_verification_2021}.
In both \cite{schwalbe_concept_2020} and \cite{rabold_expressive_2020}
a new ground truth encoding is used that highlights center points of
object part concepts. This allows to learn center point prediction
even if only binary segmentation masks are available, which is the
case for most standard CA datasets.
\cite{schwalbe_concept_2020} further suggests and compares different
optimization settings for the concept model and suggests to increase
the convolution kernel size from one pixel to a window,
in order to provide more context for each single prediction.
\CAmethod{Net2Vec for OD}{\citep{schwalbe_verification_2021}}%
This is used and further evaluated in
\cite{schwalbe_verification_2021}, where efficiency is improved to
apply a modified Net2Vec method to large object detector CNNs.
\CAmethod{TCAV}{\citep{kim_interpretability_2018}}
In parallel with Net2Vec, the similar method
TCAV~\citep{kim_interpretability_2018} for linear CAV retrieval was
developed. Instead of single pixels they consider the complete
activation map, and their linear support vector machine (SVM) concept
model does image-level concept classification.
They also suggest relative CAVs by training a concept model to
differentiate just two concepts.
Furthermore, in this work some concrete application of CAVs for
fairness analysis are suggested and demonstrated which will be
discussed later.
\CAmethod{CLM}{\citep{lucieri_explaining_2020}}
TCAV has also inspired several extensions. One of them is
CLM by \cite{lucieri_explaining_2020}, where the concept classification
is refined to a concept segmentation via the input attention to a
concept output. The intriguingly simple approach achieved mediocre
performance both for perturbation- and gradient-based heatmapping
on a generated dataset and the CelebA dataset~\citep{liu_deep_2015}.
\CAmethod{RCV}{\citep{graziani_regression_2018,graziani_concept_2020}}
Another extension of TCAV concept models are Regression Concept
Vectors (RCV) by \cite{graziani_regression_2018}.
Here, instead of binary classification, the linear model is trained to
regress a continuous-valued concept using linear least squares (LLS)
optimization. They also suggest an improved global concept-to-output
attribution score (cf.~\autoref{sec:applications.metrics}) and
demonstrate their approach on medical data in
\cite{graziani_regression_2018} and \cite{graziani_concept_2020}.
The later work \cite{graziani_concept_2020} on RCVs, just as
\cite{kazhdan_now_2020}, proposes to apply global average pooling to
the full activation maps before training the linear classifier.
This reduces the latent space and CAV size.
\CAmethod{CHAIN}{\citep{wang_chain_2020}}
Not based on TCAV or Net2Vec, but also linear in nature, are the binary
segmentation concept models proposed in the work on Concept-harmonized
HierArchical INference (CHAIN)~\citep{wang_chain_2020}.
Using Bregman iteration these are trained for a least squares distance
to concept samples.
The core idea of CHAIN is to assess the dependencies of CAVs in later
layers with respect to CAVs in earlier layers. This turns a DNN
into a concept graph of hierarchical inference.

\paragraph{Locally linear concept localization}
Some CA methods do not rely on linear concept models, but still
locally give rise to a linear interpretation and CAVs (loc.\,lin.).
\CAmethod{Local CAVs}{\citep{zhang_examining_2018}}
One such approach is pursued by \cite{zhang_examining_2018} in
order to locally represent outputs of a DNN by activation map vectors.
They consider the derivative vector of the output with
respect to the masked activation map. The mask for this purpose is
learned globally for the respective output, using a Lasso-like
optimization.
Though the global concept model---containing the derivative function of
the DNN---is highly non-linear, per-sample this still gives rise to
comparable local CAVs.
\CAmethod{AfI}{\citep{wu_global_2020}}
Another approach, which is also locally producing CAVs, is
Attacking for Interpretability (AfI)~\citep{wu_global_2020}. Here, concepts
represented in the final output classes of a classifier are located
within activation maps.
To obtain a CAV for an input instance and a selected output concept,
they consider the local neuron-to-output attribution of
neurons in a layer to the output of interest.
The local CAV is the result of a global average pooling on the
neuron attribution maps.
In this case, the attribution is not defined via sensitivity, but as the
difference between the vanilla and a perturbed version of the input,
with perturbation done by a trained global attacker model.

\paragraph{Non-linear concept localization}
\CAmethod{SeVec}{\citep{gu_semantics_2019}}
A simple clustering-based approach to obtain CAVs is followed by
SeVec in \cite{gu_semantics_2019}.
Here, the (binarized) activation vectors for positive concept examples
are clustered by cosine distance. The center of the most prominent
cluster is taken as a CAV for the concept.
From their diverse experiments, they report that there always was a
dominant cluster containing most of the positive samples. 
\CAmethod{CME}{\citep{kazhdan_now_2020}}
Another example of CAVs from clustering under supervision was used for
Concept-based Model Extraction (CME) by \cite{kazhdan_now_2020}.
Their two specialties are that they also consider multi-class concepts
like \Concept{shape} (with values, \forexample, \Concept{rectangle},
\Concept{triangle}, \Concept{circle}), and that they assume that also
unlabeled inputs are available for concept training.
Thus, they use a semi-supervised multi-task learning approach, namely
label spreading for spectral clustering\footnote{
  See, \forexample, the used implementation at
  \url{https://scikit-learn.org/stable/modules/semi_supervised.html\#label-propagation}
}, to train their concept
models. Similar performance results to Net2Vec are reported on the
dSprites dataset~\citep{matthey_dsprites_2017}
and the Caltech-USCD birds dataset~\citep{wah_caltechucsd_2011}.
\CAmethod{IIN}{\citep{esser_disentangling_2020}}
Other than the clustering-based approaches,
\cite{esser_disentangling_2020} train a bijection between a layer's
output space to a vector space of the same size. The idea is that
clusters of dimensions in that vector space represent given concepts.
The bijection is realized by invertible DNNs, so-called normalizing
flows, and termed Invertible Interpretation Network (IIN). 
An interesting label format is used: The image-level concept vectors
are trained to be the difference vector between pairs of samples
representing an increase of the concept
(\forexample, non-\Concept{smiling} $\rightarrow$ \Concept{smiling}).
They also suggest to use their approach for concept mining by only
enforcing independence of the dimension clusters.
\CAmethod{CBM student}{\citep{haselhoff_blackbox_2021}}
While all previous supervised methods assume full white-box access to
the DNN of interest, \cite{haselhoff_blackbox_2021} break with this
assumption and investigates how to obtain concept attribution
from a black-box function.
The idea pursued in the teacher-student approach in
\cite{haselhoff_blackbox_2021} is to distill a concept bottleneck
model from a trained function.
Their student model is a concept bottleneck with a
ResNet50~\citep{he_deep_2016} back-end and a Gaussian discriminant
analysis after the concept bottleneck.
This allows a supervised concept analysis on black-box functions.

\subsubsection{Concept mining}\label{sec:methods.posthoc.mining}
\methodtab{methods.posthoc.mining}
In literature, two method classes for finding concepts in latent
space representations are prominent: Clustering-based approaches, and
matrix factorization (matrix fact.) approaches. Examples of both are
discussed in the following.

\paragraph{Concept mining using clustering}
\CAmethod{Feature Visualization}{\citep{olah_feature_2017,nguyen_understanding_2019}}
As for concept localization, the simplest approaches of unsupervised
post-hoc concept analysis, or concept mining, are to investigate what
concepts are encoded by single units of a DNN, like filters of a
convolutional DNN. This is done in the broad spectrum of methods for
feature visualization~\citep{olah_feature_2017,nguyen_understanding_2019}
where exemplary inputs or generated prototypes help to explain the
meaning of such single units.
\CAmethod{ACE}{\citep{ghorbani_automatic_2019}}
An approach respecting distributed representations of information is
followed by Automatic Concept-based Explanations
(ACE) in \cite{ghorbani_automatic_2019} for visual concepts. They assume
that meaningful visual concepts are represented by superpixels, \idest
connected patches, in images. Thus, they cluster the latent space
representation of size-normalized super-pixels using k-means
clustering, and associate a concept with each cluster while the
centers of the clusters serve as CAVs.
\CAmethod{\citep{yeh_completenessaware_2020}}{}
The same approach is utilized by the direct successor
by \cite{yeh_completenessaware_2020}, but with additional custom
optimization criteria for the clustering. These shall ensure natural
properties of concepts like spatial proximity of similar concepts.
Also, they aim to capture a selection of concepts that covers the
complete amount of information encoded in one latent space.
\CAmethod{VRX}{\citep{ge_peek_2021}}
Another extension of ACE is the Visual Reasoning eXplanation framework
(VRX) from \cite{ge_peek_2021}.
Before applying the super-pixel segmentation to an image, they mask
out regions in an image that had little attribution to the output.
For the attention scoring they use the standard heatmapping method
Grad-CAM~\citep{selvaraju_gradcam_2017}.
\CAmethod{NBDT}{\citep{wan_nbdt_2020}}
Other than previous k-means based methods,
Neural-Backed Decision Trees (NBDT)~\citep{wan_nbdt_2020}
uses hierarchical agglomerative clustering (HAC)
to find a hierarchy of CAVs in the last DNN hidden layer.
The leaves of the hierarchical tree are concepts represented by the
DNN outputs, with their CAVs being the respective weights from the
last hidden layer.
Using HAC with respect to $L_2$ distance on those leaves,
a hierarchy of clusters is established.
As before, the cluster center points are interpreted as CAV.
For example, this hierarchy of CAVs can be used to
define a decision tree surrogate model.
\CAmethod{Explanatory Graph}{\citep{zhang_interpreting_2018}}
A totally different approach that is loosely related to hierarchical
and spectral clustering is pursued by the Explanatory
Graph~\citep{zhang_interpreting_2018} extraction method.
Their goal is to represent the inference of a CNN purely via
a hierarchical graph of part objects and their part-relation:
From small and simple concepts in earlier layers,
to large and complex concepts in later layers.
As constraints to define proper concepts, they assert that a concept
prediction must
(1) coincide with a high activation of a filter map; and
(2) be spatially consistent with predictions of its parent concept in
the successive layer.
Due to this hierarchical dependency of concepts, they always produce a
family of concept models that can only be inferred following the
hierarchical structure. These concept models are each anchored and
restricted to an image region.

\paragraph{Concept mining using matrix factorization}
\CAmethod{NCAV}{\citep{zhang_invertible_2021}}
A generalization of the previous clustering methods was
proposed with Non-negative CAVs (NCAV) in \cite{zhang_invertible_2021}.
They view k-means clustering as a special case of matrix
factorization, and use this to find pixel-wise linear concept models
for CNNs, \idest linear weights for the convolutional filters.
The idea of matrix factorization here is to find a global
matrix $P$ (of size $\text{\#channels}\;\times\;\text{\#concepts}$)
for a layer, where the columns of the matrix encode CAVs.
For each input a matrix $S$ is obtained via optimization in a way that
$S\cdot P$ approximately is the original activation map.
$S$ then consists of the pixel-wise scores for the concepts, \idest
the per-pixel outputs of the concept models.
The work compares three matrix factorization approaches, namely
k-means clustering, principal component analysis, and non-negative
matrix factorization for ReLU nets, with most interpretable concepts
found by the last one.
\CAmethod{\citep{saini_analyzing_2020}}{}
A similar approach to NCAV, also using non-negative matrix
factorization, is pursued by \cite{saini_analyzing_2020}.
The important difference is that they do not only associate
\emph{concepts} with filters (cf.~dimensionality of $P$), but also
input pixels and single neurons at once, leading to much more memory
intensive optimization, but a little more direct association of
concepts to input parts.

\subsubsection{Complete latent space disentanglement}\label{sec:methods.posthoc.latentspace}
Some of the discussed methods allow not only to replace parts of the
model but provide an (invertible) interpretable representation for a
complete layer output. This is relevant if it is the aim to capture as
much of the encoded information as possible, or to turn a model
post-hoc into a concept bottleneck model.
Some of the discussed methods are recapitulated here under this
aspect. 

Concretely, \cite{yeh_completenessaware_2020} suggest a measure for
evaluating completeness of a set of concepts: They intercept the
connection between two layers by an interpretable intermediate output
where each single dimension corresponds to one concept.
While concept models are used to connect the earlier layer with the
interpretable units, linear weights are learned to connect the interpretable
representation with the next layer.
They take the performance drop due to this interpretable
interception as a measure for the \defn{completeness} of the set of chosen
concepts.
Such an interpretable interception is also achieved by the matrix
factorization approach of NCAV.
Their idea to approximately represent the layer output as a product
$S\cdot P$ of interpretable matrices allows to directly replace the
layer output with this concept-based approximation.
And the NBDT approach inherently ensures to
capture all relevant information of the last hidden layer they
consider: They take into account all weights connecting the
last hidden with the output layer.
Note that the supervised IIN method
also achieves concept-based latent space disentanglement,
but adding residual \enquote{non-semantic} dimensions as a fallback
for encoded but not predefined concepts, to ensure invertibility.


\section{Applications}\label{sec:applications}
%
%
%

This chapter provides an overview of core applications of concept
analysis, and discusses interesting examples.
The first sub-section \ref{sec:applications.usingcaoutput}
considers applications that utilize the additional concept outputs
produced by concept analysis.
The interactive model correction method of concept
intervention is discussed in
\autoref{sec:applications.conceptintervention},
and \autoref{sec:applications.metrics} gives an overview of useful
metrics and approaches for qualitative verification based on CA.


\subsection{Using enriched DNN output}\label{sec:applications.usingcaoutput}
Several applications require access to a rich set of features in the
DNN (intermediate) output.
Concept analysis provides methods to either enrich the output after
training the model with few data samples, or directly and easily
include rich semantic outputs into the architecture.
We first discuss examples of model distillation utilizing concept
analysis, and then verification methods that rely on these rich
outputs.

\subsubsection{Model distillation}\label{sec:applications.usingcaoutput.modeldistillation}
By \defn{model distillation} we understand the approximation of the
functionality of the complete or parts of a DNN by a more
interpretable model, which can be, \forexample,
a linear one or a decision tree \citep{arrieta_explainable_2020}. 
Model distillation into an interpretable surrogate model requires
interpretable symbolic inputs for that surrogate model,
such as \Concept{color} or \Concept{age}.
However, many input domains for DNNs are non-symbolic, like images.
Post-hoc attached concept model outputs can be used to provide the
necessary symbolic input to the surrogate. This allows to replace
bottom parts of the DNN similar to the fine-tuning approach used for
the Semantic Bottleneck Models in \cite{losch_interpretability_2019}
(cf.~\autoref{fig:conceptbottlenecks}).
In the following, examples for different types of surrogate models are
summarized. An overview is given in \autoref{tab:proxytypes}.

\begin{table}
  \centering
  \footnotesize
  \caption{Overview of the presented model distillation approaches
    based on concept analysis outputs}
  \label{tab:proxytypes}
  \begin{tabular}{@{}>{\raggedright}p{.4\linewidth} l@{}}
    \toprule
\scriptsize\textbf{Distillation Method}	&	\scriptsize\textbf{Surrogate Model Type}	\\\midrule
CA \& ILP~\methodcite{rabold_expressive_2020}	&	Expressive rules	\\\midrule
CME~\methodcite{kazhdan_now_2020}	&	Logistic regression, decision tree	\\\midrule
Additive explainer~\methodcite{chen_explaining_2019}	&	General additive model	\\\midrule
Interpretable Basis Decomposition \methodcite{zhou_interpretable_2018}	&	Local linear model	\\\midrule	
NBDT~\methodcite{wan_nbdt_2020}	&	Decision tree (semantic hierarchies)	\\\midrule
Explanatory Graph~\methodcite{zhang_interpreting_2018}	&	Hierarchical graph (only part-relations)	\\\midrule
CHAIN~\methodcite{wang_chain_2020}	&	Hierarchical graph (inference dependencies)	\\\midrule
VRX \methodcite{ge_peek_2021}	&	Graph DNN	\\\midrule	
CBM student~\methodcite{haselhoff_blackbox_2021}	&	Concept bottleneck model	\\

    \bottomrule
  \end{tabular}
\end{table}

\paragraph{Rules and additive models}
\Proxymethod{CA and ILP}{\citep{rabold_expressive_2020}}
For example, in \cite{rabold_expressive_2020}, the later layers are
replaced by a set of rules learned using inductive logic programming.
Also, the concepts are not located in one layer, but for each concept
the layer with the best embedding is chosen.
%
\Proxymethod{CME}{\citep{kazhdan_now_2020}}
CME \citep{kazhdan_now_2020} also considers all layers to find the best
embedding of the predefined set of concepts.
They, however, rely on simpler surrogate models,
namely a linear one trained using logistic regression, and decision trees.
%
\Proxymethod{Additive explainer}{\citep{chen_explaining_2019}}
Another example of an additive surrogate model for a classifier is the
additive explainer model presented in \cite{chen_explaining_2019}.
As concept models they use the approach by
NetDissect \citep{bau_network_2017}, in order to associate a
predefined concept with the filter of a layer that mainly
activates for this concepts.
\cite{chen_explaining_2019} now trains a general additive model to
predict the final classification output from those concept model outputs.
For each instance, the general additive model predicts weights to
linearly combine the concept model outputs.
%
\Proxymethod{Interpretable Basis Decomposition}{\citep{zhou_interpretable_2018}}
A simple but local linear surrogate is considered in the Interpretable
Basis Decomposition method \citep{zhou_interpretable_2018}.
They decompose a latent space vector representation of an input into a
linear combination of CAVs.
This is done by means of greedy least squares basis decomposition.
Concretely, those local surrogates are then used to
provide per-concept attribution heatmaps.

\paragraph{Hierarchical models}
\Proxymethod{NBDT}{\citep{wan_nbdt_2020}}
The method NBDT \citep{wan_nbdt_2020} mines a hierarchy of CAVs,
each of which represents the center point of a (sub-)cluster.
They suggest to use this concept hierarchy to define a decision
tree: Top-down, at each node one chooses the sub-cluster for which the
representing CAV and the current activation vector have the smallest
dot-product (note: not $L_2$ distance as used for the clustering).
To ensure good fidelity of the decision tree, a fine-tuning loss is
tested that encourages the model to better sort activations into the
correct clusters.
%
\Proxymethod{Explanatory Graph}{\citep{zhang_interpreting_2018}}
Other than NBDT, which is restricted to the last layer,
the Explanatory Graph method \citep{zhang_interpreting_2018}
discussed in \autoref{sec:methods.posthoc.mining}
represents the complete model by a hierarchy of concepts.
For this, they mine concepts from different layers.
Here, the type of concepts and hierarchical relations are
restricted to object parts and part relations.
%
\Proxymethod{CHAIN}{\citep{wang_chain_2020}}
Similarly, but without that restriction, the super-vised method
CHAIN \citep{wang_chain_2020} produces a hierarchical inference graph.
CHAIN linearly models how concepts in later layers depend on
concepts in earlier layers.
For this, pixel-wise linear CA models are used that provide CAVs.
The main idea for obtaining the weights for linear dependency
in CHAIN is to lift CAVs of later concepts to earlier layers.
Weights are then trained to represent a lifted CAV as linear
combination of CAVs from that layer.

\paragraph{Complex models}
\Proxymethod{VRX}{\citep{ge_peek_2021}}
To capture and model both spatial and semantic relations between
mined visual concepts, VRX \citep{ge_peek_2021} distills a graph neural
network. As inference input, concepts are detected in the image
(associated with image patches), and
a fully-connected graph of the matched patches is created that encodes
spatial relationships and (learnable) semantic relationships of
concept instances.
%
\Proxymethod{CBM student}{\citep{haselhoff_blackbox_2021}}
An approach to distill an interpretable model from a classifier
applicable without white-box access is chosen in \cite{haselhoff_blackbox_2021}.
As a surrogate, they train a concept bottleneck model with
the part after the bottleneck being a decoder based on Gaussian
discriminant analysis. This setup quite easily allows to determine the
attribution of a concept to an output as the ratio of concept
log likelihood and output class log likelihood.

\subsubsection{Verification of symbolic properties}\label{sec:applications.usingcaoutput.verification}
Besides model distillation, another usage of rich DNN output is for
the verification of logical properties.
Logical properties must be defined on symbolic features in the DNN
(intermediate) output, which can be accessed using concept analysis.
An exemplary verification of hierarchical relations like
\enquote{\Concept{dog}s are \Concept{mammal}s} is conducted in
\cite{roychowdhury_image_2018}.
They translate first-order logic rules into continuous-valued fuzzy logic rules
that accept output scores of the DNN.
The truth values of the rules can then be tested on a test set.

\subsection{Concept intervention}\label{sec:applications.conceptintervention}
For interactive human-machine-systems, like experts interacting with
an assistant system, it may be helpful to allow an expert to correct
intermediate steps in the model decision process.
This can be done by changing the intermediate output of the model.
However, to allow helpful intervention, an expert both needs to
(1) understand the meaning of the intermediate output, as well as
(2) be able to modify it according to symbolic knowledge.
Part one is solved by using concept analysis methods.
Part two can be tackled using so-called \defn{concept intervention}.
There are two approaches that rely on different prerequisites:
(a)~Either a vector representation of the concept must be available, or
(b)~the concept model output must directly
influence later reasoning steps of the considered model.

\paragraph*{Concept model output intervention}
If the concept model output is part of the model inference process,
the expert can manually change the output of the concept model and
re-do the inference from the concept model output onwards.
This is suggested and done for CBMs in \cite{koh_concept_2020}.
CME \citep{kazhdan_now_2020} does this post-hoc for standard DNNs.
They attach concept outputs post-hoc and then train a surrogate upon
them. The surrogate decisions can now be intervened by changing
concept outputs.

\paragraph*{Concept representation intervention}
The other setting for concept intervention is that a latent space
vector is given that represents the direction towards a concept in the
latent space.
Then, the intermediate output vector in that latent space can be
modified to point more or less into the direction of the concept.
For example, projecting it to the plane orthogonal to the CAV
eliminates the concept from the intermediate output.
The modified vector can then be fed to the next layer.
This approach is, \forexample, used for the Counterfactual Explanation
Score (CES) \citep{abid_meaningfully_2021}, where the difference
between adding the concept and removing the concept is measured.
In GAN dissection \citep{bau_gan_2018}, the concept localization method
NetDissect \citep{bau_network_2017} is used to associate single
units in a generative adversarial network for image generation with
semantic concepts.
Intervention---\idest manipulation---of those unit values results in
spatially local semantic manipulation of the created image.


\subsection{Analysis of concept model properties}\label{sec:applications.metrics}
Concept analysis gives rise to some interesting qualitative and
quantitative analysis options.
These can be used to globally or locally verify a model by uncovering failure
modes \citep{zhang_examining_2018}, or allowing for manual inspection
of the rich semantics assessable via concept analysis.
In the following, we first discuss concept-driven qualitative
inspection methods enabled by CA
(\autoref{sec:applications.metrics.qualitative}),
and then quantitative measures related to CA results
(\autoref{sec:applications.metrics.quantitative}).

\subsubsection{Qualitative analysis methods}\label{sec:applications.metrics.qualitative}
Some qualitative analysis methods are enabled by CA which are
discussed in the following.

\begin{parlikedescription}
\item[Local input-to-concept attribution]
  In the Interpretable Basis Decomposition
  approach \citep{zhou_interpretable_2018}
  the attribution of input pixels to a concept is visualized.
  For this, they use a standard and interchangeable heatmapping
  method (Grad CAM by \cite{selvaraju_gradcam_2017}), and treat the
  post-hoc attached concept model outputs as regular DNN outputs.
  They consider concepts that define a local basis for the output in
  order to semantically dissect the attribution to the final output.
  In CLM \citep{lucieri_explaining_2020} also a perturbation based
  attribution method is applied and the heatmaps are used to allocate
  concept classification results to image regions.
\item[Concept prototypes]
  For manual inspection, it may be interesting to get an impression of
  how a model \enquote{perceives} a localized or mined concept.
  For this, one can either
  provide examples \citep{ghorbani_automatic_2019},
  patch examples to get a prototype \citep{ghorbani_automatic_2019},
  or optimize a candidate to obtain a prototype for the concept.
  The latter is done in \cite{kim_interpretability_2018} by applying the
  DeepDream \citep{mordvintsev_inceptionism_2015} optimization with
  respect to the concept outputs.
\end{parlikedescription}

\subsubsection{Quantitative properties}\label{sec:applications.metrics.quantitative}
The following semantically grounded metrics directly arise
from concept analysis.\\
\emph{Notation:}
We consider an exemplary DNN $f$ and a concept $c$ in layer $l$ with
CAV $w_c$. $T$ denotes a test set of samples, and $p$ a
single input sample.
The DNN sub-function up to layer $l$ is denoted $f_{\to l}$,
$f_{l\to}$ denotes the one from layer $l$ onwards, and---in case $f$
has multiple outputs---the function for the $k$th output of $f$ is
written $f^k$.

\begin{parlikedescription}
\item[Concept embedding quality]
  The following types of concept embedding quality measures are
  proposed in literature:
  \begin{itemize}
  \item As suggested in \cite{schwalbe_concept_2020}, the best
    achievable performance of a concept model for a concept can be
    interpreted as the \defn{concept embedding strength} or quality,
    \idest how well information about the concept is embedded in the
    DNN.
    Standard performance metrics are, \forexample,
    accuracy or $\mathrm{R}^2$ for binary classification \citep{kim_interpretability_2018},
    and set intersection over union for binary segmentation \citep{fong_net2vec_2018,schwalbe_verification_2021}.
  \item \defn{Misprediction overlap (MPO)}:
    In case of a family of concept models, \cite{kazhdan_now_2020}
    argues that it is important to assess the distribution of errors
    over concepts.
    They suggest to measure the misprediction overlap as the
    proportion of test samples for which at least $n$ relevant
    concepts are mispredicted. 
  \end{itemize}
\item[Concept similarity]
  As discussed in \autoref{sec:ca.taxonomy}, there are several
  approaches to measure the similarity of concept embeddings in the
  case CAVs are available.
  These usually rely on CAVs residing in the same layer, even though
  CAV-lifting can be performed to also compare concepts embedded in
  different layers as done in \cite{wang_chain_2020}.
  The embedding similarity can be compared with prior
  knowledge on the ground truth semantic similarity for verification
  purposes as done in \cite{schwalbe_verification_2021}.
  Here, one has to differentiate between globally and locally available
  CAVs:
  \begin{itemize}
  \item
    As discussed in \autoref{sec:ca.taxonomy}, global CAVs are often
    compared using one of
    \defn{cosine similarity} (\forexample, \cite{fong_net2vec_2018,schwalbe_verification_2021,kim_interpretability_2018})
    or \defn{$L_2$ distance} (\forexample, \cite{ghorbani_automatic_2019,yeh_completenessaware_2020,wang_chain_2020}).
  \item
    For the local CAVs used in \cite{zhang_examining_2018},
    the authors suggest to assess the
    \defn{distribution of concept relation values}
    given a test set $T$.
    As measure of concept similarity for CAVs in the same layer they
    consider cosine distance.
    They assume a prior distribution of values and apply
    Kullback-Leibler divergence to compare this with the measured
    distribution.
  \end{itemize}
\item[Local concept-to-output attribution]
  Given a sample, there are several suggestions on how to quantify the
  attribution of a concept embedding to an output (or another concept
  model output).
  \begin{itemize}
  \item \defn{Local TCAV score}:
    \cite{kim_interpretability_2018} measures local
    sensitivity \citep{baehrens_how_2010} of an output with respect to
    a CAV by measuring the partial derivative along the CAV:
    \begin{gather}
      \text{TCAV}(f;c\to k)(p) \coloneqq
      \triangledown f_{l\to}^k(f_{\to l}(p)) \cdot w_c
      \label{eq:tcavlocal}
    \end{gather}
  \item \defn{Counterfactual explanation score (CES)}:
    To robustify the TCAV sensitivity score,
    \cite{abid_meaningfully_2021} suggests to use an
    approximation of the direct derivative via a step-perturbation.
    Using concept intervention
    (cf.~\autoref{sec:applications.conceptintervention}),
    they add the concept of interest by moving the latent
    space vector one step into the direction of the CAV.
    For a step size $\delta$ (originally 10,000) the approximation
    is then
    \begin{gather*}
      \text{CES}(f;c\to k)(p) \coloneqq
      f_{l\to}^k(f_{\to l}(p) + \delta w_c) - f^k(p)\;.
    \end{gather*}
  \item
    The CHAIN method by \cite{wang_chain_2020} assesses the
    \defn{local hierarchical dependency} of pixel-wise CAVs in later
    layers to CAVs in earlier layers.
    They suggest to use the partial derivative of a concept output
    along the CAV of an earlier-layer concept as local
    concept-to-concept attribution.
  \item
    An approach not relying on CAVs is to find a
    \defn{linear surrogate model} based on concept outputs that
    directly provides concept-to-output attribution weights, as is
    done in a local fashion in \cite{chen_explaining_2019}.
  \end{itemize}
\item[Global concept-to-output attribution]
  Local concept-to-output attribution values can be aggregated on a
  test set $T$ to an approximate global attribution value.
  While this was first suggested in \cite{kim_interpretability_2018},
  the aggregation method was refined in later work:
  \begin{itemize}
  \item The original way to aggregate local TCAV scores as
    defined in \autoref{eq:tcavlocal} to a \defn{global TCAV score}
    was proposed in \cite{kim_interpretability_2018}.
    The metric is specific to classifiers since it relies on a
    partition of the test set $T$ into sub-sets $T_k$ of the
    respective output class $k$.
    What is measured is the proportion of samples of a class $k$ with
    positive local TCAV sensitivity score:
    \begin{gather}
      \text{TCAV}_{\text{global}}(f; c\to k) \coloneqq
      \frac{1}{\#T_k} \sum_{p\in T_k} \mathds{1}_{\text{TCAV}(f;c\to k)(p) > 0}
      \label{eq:tcavglobal}
    \end{gather}
  \item The \defn{normalized bidirectional relevance} (NBR)
    is an improved version of \autoref{eq:tcavglobal} suggested in
    \cite{graziani_regression_2018}.
    They aggregate local sensitivities as:
    (1) the mean sensitivity
    $\mu_{c\to k}=\mean_{p\in T_k}\text{TCAV}(f;c\to k)(p)$
    for an output class $k$ over samples $T_k$ of that class,
    (2) weighted by the concept model's R-squared value $\mathrm{R}_c^2$,
    and finally
    (3) normalized by the standard deviation $\sigma_{c\to k}$ of the sensitivity:
    \begin{gather}
      \SwapAboveDisplaySkip
      NBR(f;c) \coloneqq \frac
      {\mathrm{R}_c^2 \cdot \mu_{c\to k}}
      {\sigma_{c\to k}}
    \end{gather}
    This yields stronger sensitivity for accurate concept models and
    such with less output variance \citep{graziani_concept_2020}.
  \item
    The CHAIN method lifts CAVs from
    later layers to an earlier layer in order to represent
    the lifted CAV as linear combination of CAVs in this earlier
    layer.
    These \defn{global concept hierarchical inference weights}
    are further used to define a surrogate model
    (cf.~\autoref{sec:applications.usingcaoutput.modeldistillation}).
  \item
    Sometimes, one is interested in the influence of concept $c_A$ on
    concept $c_B$, however, has no concept model for $c_A$, but only
    concept samples.
    In this case, both the AfI framework \citep{wu_global_2020} and
    \cite{schwalbe_verification_2021} suggest to observe the
    discrepancy in $c_B$'s output 
    between a set of samples with $c_A$, and a set without $c_A$.
    In \cite{schwalbe_verification_2021} the discrepancy is simply
    measured as the \defn{discrepancy in performance} of $c_B$.
    AfI measures the \defn{maximum mean discrepancy (MMD)} \citep{gretton_kernel_2012},
    a standard measure to compare two distributions,
    between the output distributions of $c_B$ on the different input
    datasets.
  \end{itemize}
\item[Local concept interactions]
  One might be interested in the
  \defn{local interaction of concepts $c_A$ and $c_B$},
  \idest how the local attribution of $c_A$ to
  an output will change with changes to $c_B$.
  In case of CBMs, or if $c_A$ and $c_B$ have a CAV in the same layer,
  standard interaction analysis methods can be applied,
  such as integrated Hessians \citep{janizek_explaining_2020}.
\item[Concept embedding validity]
  As was already found in \cite{kim_interpretability_2018} and
  confirmed in \cite{rabold_expressive_2020}, especially linear
  (supervised) concept models are non-robust with respect to outliers,
  and the training may not be stable.
  Thus, measures are necessary that confirm whether a found CAV
  is a stable and meaningful solution or not:
  \begin{itemize}
  \item
    In \cite{kim_interpretability_2018} a \defn{t-test} is used to check
    whether several CAVs trained for the same concept behave
    consistently.
  \item
    In \cite{pfau_robust_2021} a \defn{hypothesis test} is used that checks
    whether the concept sensitivity is higher than that of a reference
    noise vector.
    This is also applicable to check validity of mined CAVs.
  \end{itemize}
  In \cite{rabold_expressive_2020} an ensembling approach is suggested
  to stabilize linear models.
  An invalid or instable concept embedding may unmask an
  inconsistently defined concept, or an inappropriate CA method.
\item[Concept intervention scores]
  The concept intervention used in \cite{koh_concept_2020}
  (manually changing the bottleneck outputs of CBMs)
  may serve to measure the influence of wrongly predicted concepts
  onto the final decision.
  In that work they compare the performance of the original CBM
  with CBMs with an increasing rate of concepts being intervened,
  \idest their concept bottleneck outputs being replaced with ground
  truth labels.
  They found that for most concepts the correct prediction is
  substantial for correct output of the final model
  (cf.~\cite[Fig.~4]{koh_concept_2020}).
  Concept intervention scores can also be seen as a local
  concept-to-output attribution measure that is perturbation-based and
  specific to CBMs.
\item[Concept completeness]
  The notion of concept completeness was introduced in
  \cite{yeh_completenessaware_2020}.
  It measures how much of the task-relevant information encoded in a
  DNN is covered by a set of concept embeddings.
  They suggest to measure this as the decrease in model performance
  when turning it post-hoc into a concept bottleneck.
  The bottleneck is inserted between layers $l$ and $l+1$ as follows:
  First, the bottleneck layer after $l$ is inserted,
  with the connections from $l$ to the bottleneck concept outputs
  being the concept models. Then, weights are learned to connect the
  bottleneck to the next layer $l+1$.
  This differs from Semantic
  Bottlenecks \citep{losch_interpretability_2019}, where the complete
  part $f_{l\to}$ is replaced and retrained.
  While the original method from \cite{yeh_completenessaware_2020}
  assumes the existence of linear models defined by CAVs,
  the performance discrepancy can be measured for
  any other method that post-hoc inserts a bottleneck
  (cf.~\autoref{sec:methods.posthoc.latentspace}).
\item[Concept interpretability]
  Due to the distributed representations occurring in a DNN,
  one unit, respectively one filter in case of CNNs, may be involved
  in representing several semantic concepts.
  In the work on NetDissect, \cite{bau_network_2017}
  suggest to take the amount of these concepts as a measure for
  the interpretability of the filter.
  \cite{fong_net2vec_2018} inverted this and took sparsity of
  a CAV (\idest how many filters are needed to encode the concept)
  as a measure for the interpretability, or \defn{disentanglement}, of
  the concept representation 
  (cf.~\autoref{sec:ca.taxonomy}).
\end{parlikedescription}


\section[Image datasets for supervised concept analysis]
{Image datasets for supervised CA}\label{sec:datasets}
%
%
%

\newcommand{\datasetlink}[1]{\\{\textit{Source:} \footnotesize\url{#1}}}

In this chapter we collect and compare different datasets that so far have been
used for supervised concept analysis methods presented in \autoref{sec:methods}.
This is meant as a reference for researchers in search of suitable
datasets for evaluating CA approaches, and to demonstrate the breadth
of application fields.
A tabular overview of the datasets and their key properties is
compiled in \autoref{tab:datasets}.

Datasets are classified as follows:
We start in \autoref{sec:datasets.large} with datasets of diverse real
world images that are suitable for practical baseline evaluation of
DNNs for complex computer vision tasks.
Next, in \autoref{sec:datasets.domainspecific}, a collection of domain
specific image concept datasets are presented, that can serve for
evaluation of domain specific applications, or of CA methods on more
complex tasks.
Lastly, an overview of \enquote{toy} and simple image datasets is
given in \autoref{sec:datasets.simple} that contain few simple concept
classes like shapes, textures, or colors. These may serve for initial
experimentation or comparison.

\begin{table}
\scriptsize
\newcommand{\datasecseprule}{\secseprule{7}}
\newcommand{\datasetcluster}[1]{\datasecseprule\multicolumn{7}{@{}l@{}}{\textbf{#1}}\\*\midrule}
\newcommand{\firstdatasetcluster}[1]{\toprule\multicolumn{7}{@{}l@{}}{\textbf{#1}}\\*\midrule}
\renewcommand*{\midrule}{\cmidrule{1-7}}  
\renewcommand{\rot}[2][70]{\rotatebox{#1}{\parbox{10ex}{\scriptsize\textbf{#2}}}}
\begin{tabularx}{\linewidth}{@{}p{5.5em}@{\,} >{\footnotesize\raggedright}X@{\,} >{\raggedright}p{2em} >{\raggedright}p{6em} S[table-format=6.0] S[table-format=4.0] c >{\tiny}l@{}}
  \caption{
  Overview of image datasets used for supervised
  concept (\Concept{C}) analysis.
  Besides concept type and label type, we collected the number of
  samples, concept classes, and the number of samples per least
  dominant class value of the concept.
  Abbreviations: attr.=attributes, avg.=average, equ.\,dist.=equally distributed.
  }
  \label{tab:datasets}
  \\
  {\rot[0]{Name}}	&		&	{\rot{\Concept{C}~Label\\Type}}	&	{\rot{\Concept{C}~Types}}	&	{\rot{\#\,Samples}}	&	{\rot{\#\,\Concept{C}~Classes}}	&	{\rot{\#\,Samples\,/ \Concept{C}~Class}}	\\
  \endfirsthead
  \caption{Overview of image datasets for CA
  (continued from page~\pageref{tab:datasets})}
  \\
  {\rot[0]{Name}}	&		&	{\rot{\Concept{C}~Label\\Type}}	&	{\rot{\Concept{C}~Types}}	&	{\rot{\#\,Samples}}	&	{\rot{\#\,\Concept{C}~Classes}}	&	{\rot{\#\,Samples\,/ \Concept{C}~Class}}	\\
\datasecseprule
\endhead
\firstdatasetcluster{Large diverse datasets (\autoref{sec:datasets.large})}
ImageNet	&	\citep{deng_imagenet_2009}	&	cls,det	&	objects, parts	&	3000000 {$<$}	&	5247	&	600 avg.	\\\midrule
BRODEN	&	\citep{bau_network_2017}	&	seg,cls	&	all	&	63305	&	1197	&	10 {$\leq$}	\\\midrule
BRODEN+	&	\citep{xiao_unified_2018}	&	seg,cls	&	all	&	57095	&	1002	&	20 {$\leq$}	\\\midrule
MS~COCO	&	\citep{lin_microsoft_2014}	&	seg	&	body parts	&	118287	&	9	&	40000 {$\leq$}	\\
\datasetcluster{Domain specific datasets (\autoref{sec:datasets.domainspecific})}
LFW	&	\citep{huang_labeled_2008}	&	cls	&	gender	&	13233	&	1	&	-	\\\midrule
LFWA+	&	\citep{liu_deep_2015}	&	cls	&	facial attr.	&	13233	&	40	&	-	\\\midrule
CelebA	&	\citep{liu_deep_2015}	&	cls	&	facial attr.	&	202599	&	40	&	-	\\\midrule
FASSEG	&	\citep{khan_multiclass_2015}	&	seg	&	facial parts	&	270	&	3	&	-	\\\midrule
Picasso	&	\citep{rabold_expressive_2020}	&	seg	&	facial parts	&	452	&	3	&	452	\\\datasecseprule
Knee X-Rays	&	\citep{koh_concept_2020}	&	cls	&	clinical attr.	&	36369	&	10	&	5\% {$\leq$}	\\\midrule
CUB	&	\citep{wah_caltechucsd_2011}	&	cls,det	&	bird attr./parts	&	11788	&	327	&	100\%	\\
\datasetcluster{Simple datasets (\autoref{sec:datasets.simple})}
GTSRB	&	\citep{stallkamp_german_2011,schwalbe_concept_2020}	&	seg	&	color, shape, part	&	50000 {$<$}	&	10	&	211 {$\leq$}	\\\midrule
A-GTSRB	&	\citep{stallkamp_german_2011,kronenberger_dependency_2019}	&	cls	&	color, shape, part	&	70000 {$<$}	&	32	&	211 {$\leq$}	\\\midrule
FMD	&	\citep{sharan_accuracy_2014}	&	cls	&	material	&	1000	&	10	&	100	\\\midrule
Google-512	&	\citep{schauerte_google512_2010}	&	cls	&	color	&	5632	&	11	&	512	\\\midrule
dSprites	&	\citep{matthey_dsprites_2017}	&	cls,det	&	shape, spatial	&	737280	&	4	&	equ.\,dist.	\\\midrule
3dshapes	&	\citep{deepmind_3dshapes_2021}	&	cls	&	shape, spatial	&	480000	&	6	&	equ.\,dist.	\\\midrule
SCDB	&	\citep{lucieri_explaining_2020}	&	seg	&	shape	&	6000	&	10	&	equ.\,dist.	\\

  \bottomrule
\end{tabularx}
\end{table}

\subsection{Large and diverse real-world image datasets}\label{sec:datasets.large}
The following large real-world image datasets with concept annotations
are used throughout several papers to establish a performance baseline
for CA
approaches \citep{fong_net2vec_2018,kim_interpretability_2018,schwalbe_verification_2021}.
They are suitable for practical evaluation of large DNNs in complex
computer vision applications like object
detection \citep{schwalbe_verification_2021}.

\begin{parlikedescription}
\item[ImageNet]
  ImageNet \citep{deng_imagenet_2009} is a popular and large image
  dataset that aims to provide sample images for the abundance of
  \num{80000} hierarchical lexical synonym sets defined in the WordNet
  lexical database \citep{fellbaum_wordnet_1998}.
  ImageNet features more than 3.2 million samples with binary
  classification labels for \num{5247} hierarchically related synonym sets.
  The number of samples per synonym set varies a lot, with an average
  of 600 samples per set, and 50\% of the sets containing more than
  500 samples.
  Selected concept classes served, \forexample, for validation purposes
  in the TCAV paper by \cite{kim_interpretability_2018}.
  \datasetlink{https://image-net.org/}
\item[BRODEN]
  The BRoadly and DEnsely labeled dataset (BRODEN) was first
  introduced in \cite{bau_network_2017}.
  It unifies diverse other image datasets
  (ADE20k by \cite{zhou_scene_2017},
  OpenSurfaces by \cite{bell_intrinsic_2014},
  Pascal-Context by \cite{mottaghi_role_2014},
  Pascal-Part by \cite{chen_detect_2014}, and
  the Describable Textures Dataset by \cite{cimpoi_describing_2014}).
  In total, Broden features \num{63305} different images resized to
  unified resolution $224\times224$, $227\times227$, or $384\times384$.
  With \num{1197} classes in total, the dataset labels encompass pixel-wise
  concept segmentations of types
  color~(11),
  texture~(47),
  material~(32),
  object~(584),
  part object~(234),
  and classification labels for 468 scene type concepts.
  The amount of labels varies greatly, going down to as few as 10
  samples per class.
  \datasetlink{https://github.com/CSAILVision/NetDissect}
\item[BRODEN+]
  In \cite{xiao_unified_2018} the original BRODEN dataset was reworked
  to make classes more distinct, and feature a sufficient number of
  samples per class to serve for DNN training. 
  Mainly, they merged highly similar concept classes,
  removed color classes and classes with too few samples,
  and renamed labels to match the naming in the Places
  dataset \citep{zhou_learning_2014}. 
  This results in \num{57095} image samples and \num{1002} classes in total.
  \datasetlink{https://github.com/CSAILVision/unifiedparsing}
\item[MS~COCO Bodyparts]
  The Microsoft Common Objects in COntext object detection dataset
  (MS~COCO) \citep{lin_microsoft_2014} features
  over 1.7 million keypoint labels
  for over \num{150000} person instances
  in \num{118287} training and \num{5000} validation images
  with comparatively high resolution.
  \cite{schwalbe_verification_2021} suggests and demonstrates a
  general method to turn person keypoint labels into CA-ready
  segmentations of body parts described by keypoints
  (here:
  \Concept{eye},
  \Concept{nose},
  \Concept{ear},
  \Concept{arm},
  \Concept{leg},
  \Concept{wrist},
  \Concept{ankle},
  \Concept{shoulder},
  \Concept{hip}).
  \datasetlink{https://cocodataset.org/}
\end{parlikedescription}

\subsection{Domain specific image datasets}\label{sec:datasets.domainspecific}
This section collects several domain specific real-world image
datasets that have so far been used for use-case specific evaluation
of CA methods.
Since the collection is dominated by concept datasets for facial
features annotated in portraits, these are gathered into their own
subsection (\autoref{sec:datasets.domainspecific.facial}) and
discussed first.
Then, two further examples of detailed labeled domain specific
datasets are presented in \autoref{sec:datasets.domainspecific.other}.

\subsubsection{Facial features concept datasets}\label{sec:datasets.domainspecific.facial}
The following facial features datasets have served for CA evaluations:

\begin{parlikedescription}
\item[LFW]
  The Labeled Faces in the Wild (LFW)
  dataset \citep{huang_labeled_2008}
  features \num{13233} portraits of \num{5749} people with labels of
  name and gender.
  It was used by \cite{kim_interpretability_2018} in the fairness
  evaluations of the TCAV method.
  \datasetlink{http://vis-www.cs.umass.edu/lfw/}
\item[LFWA+]
  The LFWA+ dataset provides the CelebA additional label types
  of 5 landmark locations and 40 binary attributes
  for the original LFW image samples.
  It was introduced in \cite{liu_deep_2015} alongside the CelebA
  dataset.
  \datasetlink{https://mmlab.ie.cuhk.edu.hk/projects/CelebA.html}
\item[CelebA]
  The Large-scale CelebFaces Attributes (CelebA)
  dataset \citep{liu_deep_2015} contains \num{202599} portraits of
  \num{10177} celebrities annotated with 5 landmark locations and
  40 different binary attributes, like
  \Concept{bald}, \Concept{mustache}, and \Concept{makeup}.
  It was used, \forexample, in \cite{lucieri_explaining_2020} for
  evaluation of the concept localization method CLM
  on a model trained on CelebA for gender classification.
  The same label classes are used as in the LFWA+ dataset.
  \datasetlink{https://mmlab.ie.cuhk.edu.hk/projects/CelebA.html}
\item[FASSEG]
  The FAce Semantic SEGmentation dataset \citep{khan_multiclass_2015}
  contains 70 frontal face and 200 multi-pose face images with
  semantic segmentation of the object part concepts
  \Concept{eye}, \Concept{nose}, \Concept{mouth}.
  The portraits are sections from images taken from the
  MIT-CBCL \citep{mitcbcldataset}, FEI \citep{thomaz_new_2010},
  and Pointing04 \citep{gourier_estimating_2004} datasets.
  FASSEG served as a basis for the generated dataset used in
  \cite{rabold_expressive_2020} for CA-based surrogate model
  creation.
  \datasetlink{http://massimomauro.github.io/FASSEG-repository/}
\item[Picasso]
  In \cite{rabold_expressive_2020} the generated Picasso dataset is
  presented for binary classification of images into
  \enquote{valid} and \enquote{invalid} face.
  In each image, clips of facial features (\Concept{eye},
  \Concept{nose}, \Concept{mouth}) from the FASSEG
  dataset \citep{khan_multiclass_2015} are pasted upon portrait images
  in which facial features were erased.
  The arrangement of the features is considered an
  \enquote{invalid} (scrambled) face if the feature arrangement is
  unnatural, \forexample, eye and nose swapped.
  For concept segmentation labels, the original segmentation
  information was retained.
  For each concept, 452 training samples and 48 test generated samples were
  used.
  \datasetlink{https://github.com/mc-lovin-mlem/concept-embeddings-and-ilp/tree/ki2020}
\end{parlikedescription}

\subsubsection{Other domain specific image datasets}\label{sec:datasets.domainspecific.other}
In the following, two other examples of domain specific image
datasets are described that feature interesting concept labels.

\begin{parlikedescription}
\item[Knee X-Rays]
  For CBM training, in \cite{koh_concept_2020} a clinical dataset of
  diagnosis from knee X-ray images was used.
  The dataset was compiled by the Osteoarthritis
  Initiative \citep{theosteoarthritisinitiative}.
  The dataset prepared by \cite{koh_concept_2020} consisted of
  \num{36369} X-ray observations of \num{4172} patients,
  at a high uniform resolution of $512\times512$ pixels.
  \cite{koh_concept_2020} filtered highly unbalanced concepts.
  Only those were included where more than 5\% of the images feature
  the non-dominant class.
  In addition to the multi-value diagnosis classification,
  binary labels for further 18 other clinical concepts are included.
  \datasetlink{https://nda.nih.gov/oai}
\item[CUB]
  The Caltech-UCSD Birds-200-2011 (CUB) dataset
  \citep{wah_caltechucsd_2011} consists of \num{11788} bird images
  with classification annotations for the 200 contained bird species,
  and additional annotations of
  15 part locations,
  312 binary attributes (\forexample, \Concept{wing color}, \Concept{beak shape}) with
  visibility information, and
  the bounding box of the depicted bird.
  The dataset was used, \forexample, for CBM training in
  \cite{koh_concept_2020}, and for concept localization in
  CME \citep{kazhdan_now_2020}.
  \datasetlink{http://www.vision.caltech.edu/visipedia/CUB-200-2011.html}
\end{parlikedescription}

In TCAV, the medical use-case of predicting diabetic retinopathy from
retinal fundus images was analyzed with respect to diagnostic concepts
like microaneurysms or pan-retinal laser scars.
However, the concept data is not freely available.

\subsection{Simple evaluation datasets}\label{sec:datasets.simple}
While large real-world datasets may be interesting for a practical
evaluation of DNNs or CA methods, simpler settings are preferred for
initial method evaluation and comparison.
Thus, the following subsection collects several small-resolution datasets that
concentrate on annotations for simple concepts like shape, color, and
texture.
The examples are classified into datasets based on real-world images
(\autoref{sec:datasets.simple.real})
and artificial ones using generated samples
(\autoref{sec:datasets.simple.artificial}).

\subsubsection{Simple real-world concept datasets}\label{sec:datasets.simple.real}
The following examples of real-world datasets provide annotations for
simple concepts and can be used to validate small computer vision DNNs
using CA.

\begin{parlikedescription}
\item[GTSRB]
  The German Traffic Signs Recognition Benchmark (GTSRB)
  dataset \citep{stallkamp_german_2011} is a classification dataset
  that features 43 classes of traffic signs, distributed
  onto more than \num{50000} small ($15\times15$ pixels) to middle
  ($250\times250$) sized real world images.
  All images are close-up and fairly frontal, and the labels provide
  the exact bounding box of the contained sign.
  The bounding box labels allow to transform images to centered and
  uniformly sized sign photo sections.
  This was utilized in \cite{schwalbe_concept_2020} to automatically
  annotate bounding boxes for contained letters, by using static
  positions.
  This served as simplistic setting for evaluating different CA
  methods.
  \datasetlink{https://benchmark.ini.rub.de/gtsrb_dataset.html}
\item[A-GTSRB]
  For the experiments in \cite{kronenberger_dependency_2019},
  an augmented version of the GTSRB~dataset was created,
  enlarging it by about 60\% and adding classification labels for
  diverse concept types:
  main color and border color (5 values each),
  shape (4), and
  contained letters (10) and symbols (13).
  The new data samples were created by domain randomization (new
  combinations of shape and color) and background randomization (adding
  patches to the background).
  \datasetlink{https://benchmark.ini.rub.de/gtsrb_dataset.html}
\item[FMD]
  The Flickr Material Database (FMD) \citep{sharan_accuracy_2014}
  provides each 100 images for ten common material classes.
  It was used alongside the Describable Textures
  Dataset \citep{cimpoi_describing_2014} contained in
  BRODEN \citep{bau_network_2017}, and
  the Google-512 dataset \citep{schauerte_google512_2010} for
  evaluation in the SeVec concept
  localization method paper \cite{gu_semantics_2019}.
  \datasetlink{https://people.csail.mit.edu/lavanya/fmd.html}
\item[Google-512]
  The Google-512 dataset \citep{schauerte_google512_2010} consists of
  512 sample object images for each of 11 basic color terms, collected using
  the Google search engine.
  It was used in \cite{gu_semantics_2019} for color concept localization.
  \datasetlink{https://cvhci.anthropomatik.kit.edu/~bschauer/datasets/google-512/}
\end{parlikedescription}

\subsubsection{Simple artificial concept datasets}\label{sec:datasets.simple.artificial}
Finally, we collect some simple artificial image datasets that
allow for first evaluations and comparisons of concept models.

\begin{parlikedescription}
\item[dSprites]
  The dSprites dataset \citep{matthey_dsprites_2017} is a popular
  simple artificial dataset for evaluation of latent space
  disentanglement methods.
  It consists of images in which a geometric shape is pasted onto
  a black canvas, with the following varying (discretized) latent
  factors:
  shape (3),
  scale (6),
  orientation (40), and
  position (each 32 values for $x$- and $y$-position).
  The dataset contains images of all \num{737280} possible
  combinations of the latent factors, in low resolution of
  $64\times64$ pixels.
  In \cite{kazhdan_now_2020} the dataset was considered for evaluation
  of concept localization and CBM methods.
  \datasetlink{https://github.com/deepmind/dsprites-dataset/}
\item[3dshapes]
  Similar to dSprites, 3dshapes \citep{deepmind_3dshapes_2021} is an
  artificial dataset generated via varying a fixed set of latent
  factors, originally used and intended for unsupervised
  disentanglement research \citep{kim_disentangling_2018}.
  The \num{480000} samples of size $64\times64$ pixels each
  depict a 3D geometric shape centered within a rectangular room, with
  varying
  floor, wall, and object color (each 10 hue values), as well as
  object scale (8),
  shape (4), and
  camera orientation (15 values).
  The dataset was used in \cite{kazhdan_disentanglement_2021} for
  comparison of CA methods with disentanglement methods.
  \datasetlink{https://github.com/deepmind/3d-shapes}
\item[SCDB]
  The synthetic Simple Concept DataBase (SCDB) for binary image
  classification was presented in \cite{lucieri_explaining_2020} in
  order to mimic challenges in skin lesion classification using
  dermatoscopic images.
  It contains \num{6000} samples for concept training with circular
  binary segmentation masks of 10 shape concepts.
  Each image contains one large geometric shape placed on black
  background, which contains and is surrounded by smaller geometric
  shapes which represent the labeled concepts.
  Class labels of the two classes are determined by simple disjunctive
  predicate rules on co-appearance of small shapes, \forexample,
  $(\text{\Concept{hexagon}}\wedge\text{\Concept{star}})
  \vee (\text{\Concept{ellipse}}\wedge\text{\Concept{star}})
  \vee (\text{\Concept{triangle}}\wedge\text{\Concept{ellipse}}\wedge\text{\Concept{star}})$.
  The position, rotation, color, count, as well as appearance of two
  task unrelated small shapes are randomized.
  \datasetlink{https://github.com/adriano-lucieri/SCDB}
\end{parlikedescription}


\section{Challenges and research directions}\label{sec:challenges}
%
%
%

The previous chapters provided a broad overview of investigated
methods, applications, and datasets.
On this basis, this section gives some possible further research
directions to enrich the field of concept analysis, and foster its
practical application.

\paragraph*{Method combinations} 
As can be seen from \methodtabsref, not all combinations of properties
for concept localization are fully leveraged so far.
Further non-linear models like clustering may be promising for supervised concept analysis, and
unsupervised concept analysis methods are still scarce,
despite their value for qualitative explanations.
Also, further investigation of mining segmentation or detection
concept models may be valuable.
And finally, further investigation in detection approaches may be promising to enable less texture-focused localization:
Other than segmentation, they use more concept information (more than one activation map pixel).

\paragraph*{Linear models and clustering}
Concept localization using linear models suffers from inherent \emph{instability}:
The hyperplane defining the concept model strongly depends on the
support points. A solution may be ensembling of models
\citep{rabold_expressive_2020}.
Furthermore, \cite{goyal_explaining_2019} showed that linear models
like TCAV \citep{kim_interpretability_2018} are mostly
\emph{overconfident} and may require additional confidence
calibration.
%
For concept mining approaches, a big challenge is to find
\emph{suitable concept candidates}.
The mined concepts should both be relevant to the
model functioning, and meaningful to humans.
The current superpixeling approaches used for clustering approaches
may not capture all types of concepts, and in general need not to be well
align with human intuition on a concept.
Moreover, other methods like matrix factorization cannot guarantee
meaningfulness of the obtained concepts \citep{zhang_invertible_2021}.
Hence, for concept mining, further priors and constraints could help
to improve meaningfulness for humans. To give an example, one could
use tracking information from temporal sequences to find independently
moving sub-parts of objects.

\paragraph*{Completeness and minimality}
When using concepts to explain the inner workings of a model as in
\cite{rabold_expressive_2020}, it is desirable to have a set of
concepts that is \emph{both minimal and complete}.
It is unclear how to achieve completeness in supervised setting
(what concepts are still missing?).
And minimality is likewise hard to achieve because solutions for
picking a generating set of concepts are not unique with both concepts
and their CAVs not being independent.

\paragraph*{Other domains and architectures}
Finally, it would be an interesting challenge to apply concept analysis
to other domains and to other DNN architectures than
standard single-frame visual tasks. Examples may be video processing,
audio or text processing with transformers.
Concepts for text and audio might be emotions, single words or typical
sentence constructions.
Also, the networks analyzed are usually small and far from sizes
needed for some important applications like automated driving
perception.
Only two of the presented methods consider complex tasks like object
detection: CSPP \citep{feifel_reevaluating_2021} and
\cite{schwalbe_verification_2021}, the latter uncovering severe
performance issues for extension to larger models in the baseline
method Net2Vec \citep{fong_net2vec_2018}.

\paragraph*{Assessment applications}
As shown in \cite{schwalbe_verification_2021}, concept analysis is
suitable for diverse verification methods applicable in domains where
responsible AI, and thus thorough functional assessment, is relevant.
It promises to both provide semantic alignment of DNN intermediate
outputs and quantitative measures that allow to define local and
global performance indicators, easing automatic assessments.
More practical evaluations could be a great guideline to improve
upon the suggested metrics, find meaningful reference values, and
finally fill gaps in current DNN assessment recommendations.

\paragraph*{Availability of data}
A challenge common to all types of supervised concept analysis:
Richly and densely annotated data is required like in the combined
Broden dataset from \cite{bau_network_2017}, which means high labeling
effort.


\section{Conclusion}
%
%
%

After a rapid development over the last years, concept embedding analysis
has matured to an interesting sub-field of explainable artificial
intelligence.
%
This survey has established a common definition of efforts and related
terms. Specifically, CA is roughly summarized as associating semantic,
human interpretable concepts to intermediate output representations of
deep neural networks (DNN), by means of simple helper concept models.
To allow comparison of related approaches, a taxonomy is suggested
that, amongst others, differentiates the types, inputs, outputs, and
supervision of concept models.

Using this scheme, more than 30 divers CA methods are reviewed,
categorized, and compared, providing a broad overview of approaches
for CA. This should provide a good starting point for researchers
seeking to position their CA related methods, or looking for ones that
suit their use-case.
%
Applications and use-cases are also gathered and discussed in detail.
The most often encountered one is model distillation using concept outputs.
Nevertheless, other interesting applications are found and reviewed,
like intervention of concept outputs for interactive
human-machine-systems, and verification relying either on the
additional concept outputs, qualitative assessments, or metrics arising
from CA results.
An extensive list of metrics occurring in literature is
compiled that allows to easily find appropriate measures for DNN
assessment using CA.
%
For the practical entrance to the topic, a broad collection of
image datasets with concept labels useful for evaluation of CA methods
is provided. Datasets are classified according to their use-cases in
CA research. Relevant statistics and traits are summarized,
showing that many helpful resources are currently available.

Altogether, this review should give researchers interested in
applications or methods for semantic DNN latent space assessment a
good starting point, and clearly establish the XAI sub-field of
concept embedding analysis.
We look forward to more interesting results in the
field, especially with respect to the identified open research
challenges:
more datasets;
improvement of the linear and unsupervised clustering approaches;
selecting a good set of concepts for explaining a DNN (especially one
that is both minimal and complete);
and leveraging CA in further practical and complex use-cases and
applications.



%
%
%

\section*{Declarations}

\subsection*{Funding}
\begin{parlikedescription}
\item[Funding]
  The research leading to these results was partly funded by by the
  German Federal Ministry for Economic Affairs and Energy within the
  projects
  \enquote{KI Wissen – Automotive AI powered by Knowledge}
  and \enquote{KI Absicherung – Safe AI for automated driving}.
  Thanks to the consortia for the successful cooperation.
\item[Employment]
  This work was authored in the course of doctoral research in an
  employment at Continental Automotive GmbH, Germany.
\end{parlikedescription}

\subsection*{Competing interests}
\begin{parlikedescription}
\item[Non-financial interests]
  This doctoral research was supervised by Prof.\,Dr.\,Ute Schmid at the
  University of Bamberg.
\item[Financial interests]
  The author declares that they have no financial interests.
\end{parlikedescription}

\subsection*{Ethics approval} Not applicable.
\subsection*{Consent to participate} Not applicable.
\subsection*{Availability of data and materials} Not applicable.
\subsection*{Code availability} Not applicable.
\subsection*{Authors’ Contribution Statement} No contributors were involved other than the single author.


\bibliography{literature}

\end{document}